\documentclass[10pt,journal,compsoc]{IEEEtran}
\usepackage{ifpdf}
 \ifpdf
 \else
 \fi

\ifCLASSOPTIONcompsoc
  \usepackage[nocompress]{cite}
\else
  \usepackage{cite}
\fi
\ifCLASSINFOpdf
   \usepackage[pdftex]{graphicx}
   \graphicspath{{../pdf/}{../jpeg/}}
   \DeclareGraphicsExtensions{.pdf,.jpeg,.png}
\else
   \usepackage[dvips]{graphicx}
   \graphicspath{{../eps/}}
   \DeclareGraphicsExtensions{.eps}
\fi
\usepackage{amsmath}
\usepackage{algorithm}
\usepackage{algpseudocode}
\usepackage{lscape}
\usepackage{microtype}                 
\PassOptionsToPackage{warn}{textcomp}  
\usepackage{textcomp}                  
\usepackage{mathptmx}                  
\usepackage{tabu}                      
\usepackage{booktabs}                  
\usepackage{makecell}
\usepackage{color}

\usepackage[normalsize]{subfigure}
\usepackage{amsfonts}

\usepackage{amsmath}
\DeclareMathOperator*{\argmax}{arg\,max}

\usepackage{array}

\ifCLASSOPTIONcompsoc
  \usepackage[caption=false,font=footnotesize,labelfont=sf,textfont=sf]{subfig}
\else
  \usepackage[caption=false,font=footnotesize]{subfig}
\fi

\begin{document}
%
\title{voxel2vec: A Natural Language Processing Approach to Learning Distributed Representations for Scientific Data}
%
%
%
%

\author{Xiangyang He, Yubo Tao, Shuoliu Yang, Haoran Dai, and Hai Lin
\IEEEcompsocitemizethanks{
\IEEEcompsocthanksitem Xiangyang He is with State Key Laboratory of CAD\&CG, Zhejiang University. E-mail: xiangyanghe@zju.edu.com.
\IEEEcompsocthanksitem Yubo Tao (corresponding author) is with State Key Laboratory of CAD\&CG, Zhejiang University. E-mail: taoyubo@cad.zju.edu.cn.
\IEEEcompsocthanksitem Shuoliu Yang is with State Key Laboratory of CAD\&CG, Zhejiang University. E-mail: lucida@zju.edu.cn.
\IEEEcompsocthanksitem Haoran Dai is with State Key Laboratory of CAD\&CG, Zhejiang University. E-mail: haoran.dai@zju.edu.cn.
\IEEEcompsocthanksitem Hai Lin (corresponding author) is with State Key Laboratory of CAD\&CG, Zhejiang University. E-mail: lin@cad.zju.edu.cn.
}}

%
%

\markboth{IEEE TRANSACTIONS ON VISUALIZATION AND COMPUTER GRAPHICS}%
{Shell \MakeLowercase{\textit{et al.}}: Bare Demo of IEEEtran.cls for Computer Society Journals}
%



\IEEEtitleabstractindextext{%
\begin{abstract}

Relationships in scientific data, such as the numerical and spatial distribution relations of features in univariate data, the scalar-value combinations' relations in multivariate data, and the association of volumes in time-varying and ensemble data, are intricate and complex. This paper presents voxel2vec, a novel unsupervised representation learning model, which is used to learn distributed representations of scalar values/scalar-value combinations in a low-dimensional vector space. Its basic assumption is that if two scalar values/scalar-value combinations have similar contexts, they usually have high similarity in terms of features. By representing scalar values/scalar-value combinations as symbols, voxel2vec learns the similarity between them in the context of spatial distribution and then allows us to explore the overall association between volumes by transfer prediction. We demonstrate the usefulness and effectiveness of voxel2vec by comparing it with the isosurface similarity map of univariate data and applying the learned distributed representations to feature classification for multivariate data and to association analysis for time-varying and ensemble data.

\end{abstract}

}

\maketitle

\IEEEdisplaynontitleabstractindextext

%
\IEEEpeerreviewmaketitle

\IEEEraisesectionheading{\section{Introduction}\label{sec:introduction}}

\IEEEPARstart{T}{o} describe complex physical phenomena, large-scale scientific simulations such as those for computational fluid dynamics and meteorological simulation generally produce time-varying and multivariate scientific data in many fields. 
A simulation can also generate ensemble data with different parameters to simulate the phenomena under different conditions~\cite{BURCKNER:2010:RDE}. For example, in the deep water impact ensemble simulation, an asteroid is an object that is injected into the water. It is characterized by different attributes such as velocity, temperature, and pressure with different parameters over time. It is necessary to classify features in the data and explore the association between volumes to help domain experts effectively understand such complex phenomena. This brings about the requirements of effective feature extraction and association analysis methods through investigating volumes and their relationships for volume analysis and visualization.

Unlike features in univariate data, features in multivariate data are often defined by different scalar-value combinations of multiple variables in different spatial regions~\cite{GUO:2012:SMV}~\cite{LIU:2015:AAF}~\cite{LU:2017:MVD}. Associations between these fields are also value-dependent and space-dependent. Therefore, it is difficult or even impossible to extract important features using one single field for a complex physical phenomenon.
Recently, a number of interactive methods have been proposed to specify scalar-value combinations in multivariate transfer functions, such as the parallel coordinate~\cite{Inselberg:1985:TPW}~\cite{Guo:2011:MTF}~\cite{GUO:2012:SMV} and the scatter plot matrix~\cite{LU:2017:MVD}. These interactive methods are flexible when used for extracting various features, 
but it is still tedious and challenging to extract meaningful features for users who have little prior knowledge. This is because of the vast exploration space of scalar-value combinations. As reported by Parsons et al.~\cite{PAR:2004:SCF}, most valuable features are usually associated with specific variables and scalar values. Thus, it is desirable to introduce effective analysis methods to achieve automatic or semi-automatic feature classification for multivariate data.

There are different analysis tasks for time-varying and ensemble data. These include key time steps selection~\cite{ZHOU:2018:KTS}~\cite{PORTER:2019:ADL}, feature tracking~\cite{WANG:2013:SPF}, and feature evolution~\cite{LUKASCZYK:2017:NTG} ~\cite{LUKASCZYK:2019:DNT} for time-varying data, and exploration in $3$D space~\cite{LEISTIKOW:2019:AEV} and parameter space~\cite{BURCKNER:2010:RDE} for ensemble data.
Many analysis tasks depend on the overall association between volumes, including numerical association and spatial association. Numerical association mainly measures the overall scalar value association between volumes. Common methods include correlation coefficient and mutual information. Correlation coefficient can detect linear dependencies under the assumption of normally distributed data between volumes~\cite{SAUBER:2006:MAA}~\cite{SUKHAREV:2009:CSO}, while mutual information is based on entropy, in order to measure more general (non-linear) dependencies~\cite{BISWAS:2013:AIF}~\cite{DUTTA:2017:PIG}. Spatial association is mainly based on spatial structures, to estimate the association between voxels~\cite{SAUBER:2006:MAA}~\cite{NAGARAJ:2011:AGC}, or to estimate topological structures~\cite{SCHNEIDER:2008:ICO}~\cite{SCHNEIDER:2013:ICO}~\cite{CARR:2014:JCN}. 
However, little work has been done to combine numerical and spatial distributions together. In this paper, the context distributions of each scalar value are encoded as a distributed representation, which combines the two kinds of distributions to explore associations between scalar values in the same volume and in different volumes.

Isosurface-based similarity~\cite{BRUCKNER:2010:ISM} 
applies the spatial proximity of isosurfaces to measure the similarity between scalar values and then presents the similarity in an isosurface similarity map (ISM). Inspired by this method, we propose a novel unsupervised representation learning method, named voxel2vec, based on scalar values' spatial proximity to generate a low-dimensional vector, i.e., the distributed representation, as the intermediate representation for each scalar value.
We consider the scalar value of each voxel as a symbol, e.g., a word. Note that symbols will be interpreted as scalar-value combinations for multivariate data. The neighboring voxels' scalar values of the central voxel are the context of the scalar value (symbol) associated with the central voxel. In representation learning~\cite{BENGIO:2013:RLA}, if two symbols have similar contexts, they are semantically similar, such as synonyms in natural languages. In volume data, if two scalar values have similar contexts, they usually have high similarity and belong to the same feature. Similar to word embeddings~\cite{COLLOBERT:2011:NLP}, we propose to apply the skip-gram model~\cite{MIKOLOV:2013:EEO}~\cite{MIKOLOV:2013:DRO} with negative sampling to learn the distributed representation for each scalar value. The similarity between scalar values is directly quantified by the cosine similarity of the learned distributed representations. Frequency-based negative sampling in word embedding is used to improve training efficiency~\cite{COLLOBERT:2011:NLP}. 
When this method is directly applied to volume data, some positive samples may be sampled as negative samples due to the lower number of scalar values compared with words. Therefore, we modify the negative sampling strategy to improve the accuracy of voxel2vec.

In this paper, we propose voxel2vec to represent scalar values as low-dimensional vectors to preserve the spatial proximity between them and then use the vectors to quantify the similarity between scalar values. For univariate data, the spatial distribution of one scalar value is its isosurface; the spatial proximity between scalar values is equivalent to the similarity of isosurfaces. Thus, voxel2vec can generate a similarity map similar to the ISM. For multivariable data, we regard the scalar-value combinations as symbols, extend voxel2vec to learn the distributed representations of scalar-value combinations, and implement feature classification by high-dimensional clustering. For time-varying and ensemble data, the volumes corresponding to adjacent time step/ensemble parameters have certain continuity; therefore, we propose a transfer prediction method, that is, predicting the scalar values of central voxels of other volumes using the trained model, to achieve quantification of the association between volumes.

To summarize, the contributions of this paper are as follows:
\begin{itemize}
	\item An unsupervised representation learning model to automatically learn context distributed representations of scalar values/scalar-value combinations in volume data for measuring the relationships between them.
	\item Feature classification based on distributed representations of scalar-value combinations for multivariate data.
	\item Association analysis based on transfer prediction for time-varying and ensemble data.
\end{itemize}

\section{Related Work}

Our research addresses the application of representation learning in volume data. Volume data analysis and visualization are key topics in scientific visualization, in particular, feature classification and association analysis. We refer our readers to the survey papers~\cite{FUCHS:2009:VOM}~\cite{Wang:2019:VAV}~\cite{HE:2019:MSD}~\cite{BAI:2020:TVV} for an overview. In this section, we review some of the most closely related work on feature classification for multivariate data and association analysis for time-varying and ensemble data and summarize the application of representation learning in visualization.

\subsection{Feature Classification}

Interactive feature classification methods allow users to specify features of interest in high-dimensional transfer functions. Zhao and Kaufman~\cite{ZHAO:2010:MRA} introduced parallel coordinates to present multiple variables and their scalar value distributions. Features can be classified by specifying the ranges of related variables. Guo et al.~\cite{GUO:2012:SMV} further extended parallel coordinates with scatter plots based on multidimensional scaling (MDS)~\cite{TORGERSON:1952:MSI} and classified features based on both the advantages of parallel coordinates and scatter plots without switching the context. Lu and Shen~\cite{LU:2017:MVD} employed a scatter plot matrix to inspect all single and two-variable subspaces as well as provided a bottom-up exploration framework for interactively extracting and modifying features. These high-dimensional transfer functions require guidance to facilitate user exploration. Otherwise, it would be time-consuming and less effective to classify features in the huge feature space.

Many data mining methods have been used to simplify the feature classification process by firstly automatically analyzing data and then interactively exploring it. Wu et al.~\cite{WU:2015:WAF} applied the statistical region merging algorithm to cluster features. These features can be interactively adjusted using a suite of dimension reduction, visual encoding, and filtering widgets. Zhou and Hansen~\cite{Zhou:2014:GSS} classified features through an automatic feature extraction approach based on the slice-guided uncertainty-aware lassos. Wang et al.~\cite{WANG:2015:MPM} presented a pattern-matching approach for multivariate time-varying data that allows users to define a set of 3D scale-invariant feature transform (SIFT) in multiple fields and then automatically locate and rank matching patterns over all time steps. He et al.~\cite{HE:2018:ACF} employed a biclustering algorithm to simultaneously group variables and voxels to generate feature subspaces, i.e., voxels with a similar scalar-value pattern over a subset of variables. Users can interactively explore feature subspaces to identify features of interest. Our method also belongs to this category, in which features are first automatically clustered based on distributed representations of scalar-value combinations, and users can then interactively explore and adjust these features. Generally, previous methods have directly clustered voxels based on scalar values. In contrast, our method derives intermediate representations of voxels to classify features more effectively.

\subsection{Association Analysis}

Volume data are usually multivariate, spatiotemporal, and even ensemble. Therefore, the relationships among volumes are generally hidden and complex.

Researchers have used visualization methods widely to evaluate the inherent association and uncertainties in ensemble data. Most work focuses on the visual analysis of the association between volumes. Biswas et al.~\cite{BISWAS:2016:VOT} investigated the association between parameter sensitivity and spatial resolution accuracy based on the Weather Research and Forecasting model under different parameters and resolutions. Ferstl et al.~\cite{FERSTL:2016:TCA} focused on the association of weather and climate data and analyzed the time evolution and variability. Ma and Entezari~\cite{MA:2018:AIF} used a high-density clustering algorithm to study the distribution behaviors of iso-lines in ensemble data and proposed an interactive framework for the ensemble visualization of iso-lines. Wang et al.~\cite{WANG:2016:MCE} employed parallel coordinates to understand intra-set parameter correlation. Bruckner et al.~\cite{BRUCKNER:2010:BEO} proposed a result-driven exploration approach to analyze ensemble simulations. They split each volumetric time sequence into similar parts over time and then grouped different parameter spaces using a density-based clustering algorithm. Leistikow et al.~\cite{LEISTIKOW:2019:AEV} directly analyzed the association and difference bwtween volumes under different parameters in $3D$ space. Tkachev et al.~\cite{TKACHEV:2019:LPM} predicted future scalar values at a given position based on the past values in its neighborhood and applied the prediction errors to the analysis of ensemble dissimilarity.

For the association analysis of time-varying data, Dutta et al.~\cite{DUTTA:2017:PIG} presented a unified analysis framework based on mutual information to create the time-varying pointwise mutual information (PMI) field to capture the information from multiple time steps. Tao et al.~\cite{TAO:2018:ETM} applied affinity propagation to cluster time sequences based on a temporal similarity map. However, they failed to obtain the volumes' overall similarity between time steps. Zhou and Chiang~\cite{ZHOU:2018:KTS} applied the concept of information entropy to measure the association between volumes in different time steps and then selected several key time steps with the most representative features for visualization. Obermaier et al.~\cite{OBERMAIER:2015:VTA} designed a visual analysis system dedicated to identifying trend features and abnormal members in time-varying ensemble data to explore the parameter spaces and temporal features. Han et al.~\cite{HAN:2020:V2V} presented a deep learning framework to learn the representations of transferable variables and synthesize variable sequences of multivaraiate time-varying data.

As described in~\cite{KEHRER:2012:VAV} and~\cite{Wang:2019:VAV}, advanced data abstraction and aggregation techniques are required to detect data trends and outliers, and previous methods are not effective enough to provide a criterion for association analysis of the time sequences and parameter spaces for time-varying and ensemble data. Therefore, we use voxel2vec to learn the distributed representations of scalar values and adopt transfer prediction to measure the overall association of volumes, and then apply it to time-varying and ensemble data.

\subsection{Representation Learning in Visualization}

Representation learning~\cite{BENGIO:2013:RLA} is an unsupervised learning technique that enables learning data representations by extracting latent information from multiple aspects; these representations are useful for downstream machine learning tasks. In the natural language process, Mikolov et al.~\cite{MIKOLOV:2013:EEO}~\cite{MIKOLOV:2013:DRO} proposed a two-layer neural network language model named word2vec with the skip-gram and continuous bag of words (CBOW) frameworks to learn the distributed representation for each word, which are the basis of most deep learning-based applications. In network analysis, node embeddings~\cite{PEROZZI:2014:DOL}~\cite{GROVER:2016:NSF} learn to represent nodes as low-dimensional vectors to preserve the proximity between nodes, and are used for node classification, link prediction, and anomaly detection.

In recent years, representation learning has been applied to visualization. Berger et al.~\cite{BERGER:2016:CCD} presented cite2vec to learn embedding vectors for both words and documents for exploring and detecting document collections by their citation contexts. Zhou et al.~\cite{ZHOU:2018:VAO} used the word2vec model to represent Origin-Destination streams as vectors and facilitate the exploration of spatiotemporal data. Xu et al.~\cite{XU:2018:EEO} proposed diachronic node embeddings to discover communities with similar structural proximity and temporal evolution patterns of dynamic networks. In scientific visualization, Han et al.~\cite{HAN:2018:FAD} applied an autoencoder to learn the representations of streamlines and stream surfaces for interactive clustering and selection. Further, Porter et al.~\cite{PORTER:2019:ADL} applied an autoencoder for learning the latent representation of each volume in time-varying multivariate data for representative time step selection. Our work is along the same lines as the above-mentioned research. We propose voxel2vec in order to learn the latent representations of scalar values/scalar-value combinations in volume data and to quantify the relationships between scalar values in univariate data, scalar-value combinations in multivariate data, and associations between volumes in time-varying and ensemble data.

\begin{figure}[t]
	\centering
	\includegraphics[width=0.97\columnwidth]{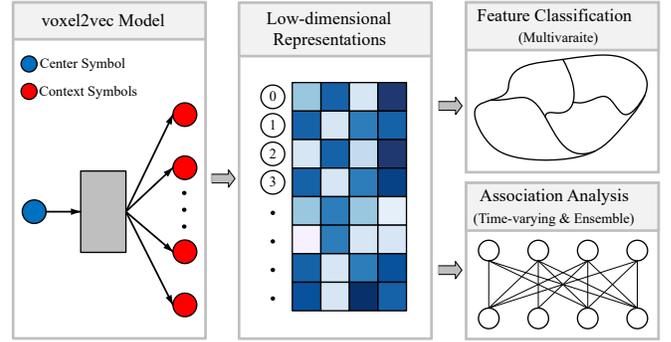}\\
	\caption{
		An overview of voxel2vec for distributed representation learning of scalar values and its applications.
	}\label{fig:basic_framework}
\end{figure}

\section{voxel2vec}
\label{sec:algorithm}

voxel2vec represents scalar values/scalar-value combinations in volume data as distributed representations to capture complex relationships between them, in particular, semantic similarity. In natural language processing (NLP), a neural language model can embed words in a low-dimensional vector space to capture both the syntactic and semantic information of words. The skip-gram model with negative sampling~\cite{MIKOLOV:2013:DRO} is a framework for learning word embeddings, and has been extensively used in many downstream NLP tasks, such as word analogy~\cite{BOJANOWSKI:2017:EWV}, named entity recognition~\cite{ZIRIKLY:2015:NER}, text classification~\cite{LAI:2015:RCN}, and machine translation~\cite{ZHANG:2014:BPE}. We apply the skip-gram model with negative sampling to capture the similarity of scalar values in a low-dimensional vector space. This representation can be used for feature classification and association analysis, as illustrated in Fig.~\ref{fig:basic_framework}. In this section, we describe data discretization and the skip-gram model, present how to modify negative sampling in the context of volume data to improve performance, and finally illustrate the learning process.

\subsection{Data Discretization}

We represent each scalar value/scalar-value combination as a symbol, i.e., a word.
voxel2vec can not be applied to the volumes in a 32-bit float format directly, since so many scalar values may lead to the model not having enough training samples for each scalar value/scalar-value combination. Similarly, each word appears in the text only once, and there would be only one positive sample for each word in natural language. Therefore, it is necessary to discretize scalar values to ensure sufficient training samples for each scalar value and capture common context information.

Volumes may have already quantized scalar values, such as most CT volumes. For volumes with continuous scalar values, we define a quantization level $R$ to reduce the number of scalar values to be learned. After discretization, each scalar value represents a small range of scalar values in the original volumes. For univariate data,  each scalar value can be directly represented as a symbol, such as the number 25. The number of distinct scalar values ($R$) is lower than the number of voxels in one quantized volume, because many voxels have the same scalar value. Given the set of scalar values $C$, voxel2vec learns one vector $z_c \in \mathbb{R}^d$ for each scalar value $c \in C$, where $d$ is the dimension of vector space. Since each voxel contains one scalar value/scalar-value combination, each voxel can be represented as one vector.

\subsection{Skip-gram Model}

\begin{figure}[t]
  \centering
  \includegraphics[width=0.9\columnwidth]{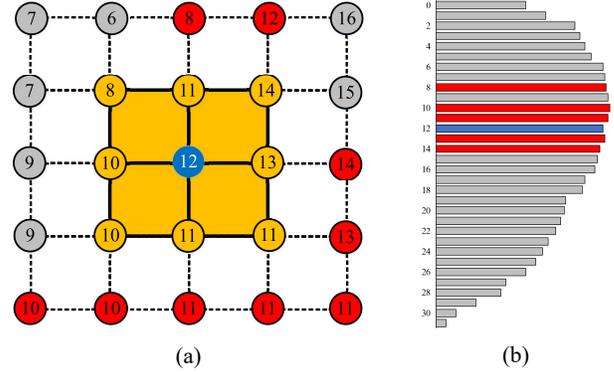}\\
  \caption{An exmaple of negative sampling in a 2D scalar field. (a) The blue voxel in the scalar field is the current central voxel with the scalar value $c_t=12$, and its neighboring voxels in the orange context window ($n=1$) become the context scalar value set $O_t = \{8, 10, 11, 13, 14\}$. The voxel outside the context window is labeled red if its scalar value is in $\{c_t \cup O_t\}$. (b) The histogram of scalar values. The bar is red if the scalar value is in $\{c_t \cup O_t\}$. The popular neighbor problem occurs if negative samples are obtained from the distribution.
  }\label{fig:optimization_2}
\end{figure}

The skip-gram model~\cite{MIKOLOV:2013:DRO} in NLP predicts co-occurrence relationships between words using the central word to predict context words in a large body of text. The training set is a sequence of sentences, and context words are the words before and after the central word in a sentence. The goal of the skip-gram model not only maximizes the co-occurrence probability of word pairs in the training set, but also minimizes the co-occurrence probability of any other word pairs that do not occur in the training set. voxel2vec regards the symbols of scalar values as words, and all scalar values in the volume are the training set. For each voxel, the scalar values of its neighboring voxels are a context of the scalar value of the voxel. 
Unlike the original skip-gram model~\cite{MIKOLOV:2013:DRO} using a 1D-window to obtain context words, voxel2vec uses a 3D-window to obtain positive samples with spatial context information. 
The \textit{context-window size} for each dimension is denoted as $n$, and the number of scalar values in the context is $(2n+1)^3-1$. Fig.~\ref{fig:optimization_2}(a) shows the eight context voxels (in yellow) of the central voxel (in blue) when $n=1$ in 2D.

For each voxel in volume data, there is a central scalar value $c$ and a context (outside) scalar value set $O$. The skip-gram model predicts the context scalar values given the central scalar value and maximizes the objective function as follows:
\begin{equation}
	\label{equ:sum}
	J = \frac{1}{T} \sum_{t=1}^{T} \sum_{o_{i} \in O_t }\log P(o_{i} \mid c_t),
\end{equation}
where $T$ is the number of voxels and $P(o_i|c_t)$ is the probability of the context scalar value being $o_i$ given the central scalar value $c_t$ of the $t$-th voxel.
One pair of ($c_t$, $o_i$) is a positive sample, such as (12, 8) shown in Fig.~\ref{fig:optimization_2}(a).

For each scalar value $c \in C$, there is one distributed representation $z_c \in \mathbb{R}^d$ when $c$ is the central scalar value, and one context vector $\hat{z}_c \in \mathbb{R}^d$ when $c$ is the context scalar value. $Z \in \mathbb{R}^{d \times R}$ and $\hat{Z} \in \mathbb{R}^{d \times R}$ are the matrixes composed of the $z$ and $\hat{z}$ of all scalar values, respectively.
The cosine similarity between $z_{c_t}$ and $\hat{z}_{o_i}$ is used to calculate 
$P(o_i|c_t)$ 
as follows:
\begin{equation}
\label{equ:probablity}
\log P(o_{i} \mid c_t) = \log \frac{exp(\hat{z}_{o_i}^T  z_{c_t})}{\sum_{w\in C} exp(\hat{z}_w^T  z_{c_t})}.
\end{equation}
The normalization factor is computationally too expensive, requiring an overall summation of all scalar values. 
Negative sampling is proposed to approximate the probability calculation by randomly drawing a negative scalar value $w$ from a distribution $D$. The primary idea is to train binary logistic regression for the positive pair ($c_t$, $o_i$) versus several negative pairs ($c_t$, $w$). 
For example, one pair of ($c_t$, $w$) is (12, 7) as one negative sample in Fig.~\ref{fig:optimization_2}(a). 
Thus, Eq.~\ref{equ:probablity} can be approximated as follows:
\begin{equation}
\label{equ:negative}
\log P(o_{i} \mid c_t) = \log \sigma(\hat{z}_{o_i}^T  z_{c_t}) + \sum_{j=1}^{k} \mathbb{E}_{w \sim D} [\log \sigma(-\hat{z}_{w}^T z_{c_t})],
\end{equation}
where $k$ is the number of negative samples and $\sigma$ is the sigmoid function $\sigma(x) = 1 / (1+exp(-x))$. This approximation increases the likelihood of the central scalar value and its co-occurring contexts. Consequently, their distributed representations are close in vector space. The approximation also enables the central scalar value to be sufficiently distinct from randomly drawn contexts; therefore, their distributed representations are consequently far apart in vector space.

\subsection{Negative Sampling}\label{sec:negative_sampling}

Negative sampling replaces the summation of all scalar values with $k$ negative samples to approximate the objective function, which considerably reduces time complexity. 
In NLP, the most frequent words (e.g., ``in'', ``the'', and ``a'') usually provide less information than rare words. Similarly, the scalar values with high frequency in volumes may or may not contain more information. For example, if scalar values belong to the background, such as 0 in univariate data, they are less informative and should not be frequently sampled during training. We use the subsampling approach~\cite{MIKOLOV:2013:DRO} to solve the imbalance between the rare and frequent scalar values in negative sampling.

However, such negative sampling involves the popular neighbor problem, as proved in node embeddings~\cite{ARMANDPOUR:2019:RNS}: a node connected with a high-degree node has a higher probability of sampling its neighboring nodes as negative samples, but its neighboring nodes as the context are positive samples. As shown in Fig.~\ref{fig:optimization_2}(b), the scalar-value frequencies of neighboring voxels are higher than other scalar values, so they are more likely to be sampled as negative samples. The popular neighbor problem must be solved because it confuses samples in different contexts and may learn ambiguous distributed representations. 
Besides, negative samples are sampled from a pre-defined static distribution $D$ that does not change during training. The informativeness of each scalar value is determined by its frequency before training. However, the informativeness of each scalar value does not stay constant but instead evolves during the training process. Some scalar values are easily classified as negative samples for the current central voxel, while others are difficult to classify, leading to over-training or under-training of samples.
In this section, we adopt two strategies to improve negative sampling for voxel2vec: adpative negative sampling to solve the popular neighbor problem and self-paced learning to sample nagetive samples dynamically based on informativeness. Sec.~\ref{sec:optimization_verification} provides an ablation experiment with/without the two strategies.

\subsubsection{Strategy 1: Adaptive Negative Sampling}

Adaptive negative sampling is designed to solve the popular neighbor problem. 
In the traditional negative sampling, $D$ is the unigram distribution raised to the 3/4-rd power for all scalar values. However, context scalar values in $O_t$ should not be sampled as negative samples for the current central voxel with $c_t$. Thus, we restrict negative sampling only from $C - \{c_t \cup O_t\}$, i.e., the probabilities of scalar values in $\{c_t \cup O_t\}$ are set to zero for the current central voxel with $c_t$. 
For this distribution, 
positive samples in $\{c_t \cup O_t\}$ would be not sampled as negative samples. As the context of each voxel is different, $D$ will be adaptively adjusted for each voxel during negative sampling. As shown in Fig.~\ref{fig:optimization_2}(b), negative samples can only be drawn from scalar values in gray for the current blue voxel.

To further solve the popular neighbor problem, we use an $L_2$ penalty onto the norm of vectors~\cite{ARMANDPOUR:2019:RNS}, because norm penalization can decrease the degrees of freedom of representation learning and prevent overfitting~\cite{FRIEDMAN:2001:TEO}. Eq.~\ref{equ:negative} is then adjusted as follows:
\begin{equation}
\label{equ:negative-1}
\begin{split}
\log P(o_{i} \mid c_t) = \log \sigma(\hat{z}_{o_i}^T  z_{c_t}) - \lambda ||z_{c_t}||_2 + \\
\sum_{j=1}^{k} \mathbb{E}_{w \sim D} [\log \sigma(-\hat{z}_{w}^T z_{c_t}) - \frac {\lambda}{k+1} ||\hat{z}_w||_2],
\end{split}
\end{equation}
where $\lambda$ is the pre-determined penalty coefficient based on the careful estimation of the derivative of the whole objective function. In the experiment, $\lambda$ is set to 0.005. The hyper-parameter analysis of $\lambda$ can be found in the supplementary material. Norm penalization provides a more stable solution for voxel2vec.

\subsubsection{Strategy 2: Self-Paced Learning}

The manner of selecting training samples for a good model has been an important research topic in the machine learning community~\cite{BUDA:2018:ASS}. To mimic the human cognitive activity of learning starting from easy concepts and moving to difficult ones, self-paced learning~\cite{KUMAR:2010:SLF} has been proposed such that training samples could be voluntarily selected by measuring the performance of samples. Gao and Huang~\cite{GAO:2018:SNE} have applied self-paced learning to negative sampling for node embeddings to achieve better prediction accuracy on node classification and link prediction. Self-paced learning is thus adopted for voxel2vec to perform negative sampling from easy samples to dynamically hard ones. Learning easy samples first and then the difficult ones makes it easier to train a better solution for the model.

Similar to adaptive negative sampling, the distribution not only depends on the current scalar value but also evolves with the current informativeness of scalar value during the training process. For the current central voxel with $c_t$, the value or informativeness of the negative sample $w \in C$ is defined as $p(w \mid c_t) = \sigma (\hat{z}_w^T z_{c_t})$. If $p(w \mid c_t)$ is large, $w$ is a difficult negative sample; otherwise, it is an easy negative sample.
To optimize the selection of negative samples, we define the distribution of drawn negative samples based on the informativeness as follows:
\begin{equation}
\label{equ:distribution}
p_{c_t,c_i} = \frac{exp(\hat{z}_{c_i}^T z_{c_t})}{\sum_{j=1}^{k}exp(\hat{z}_{c_j}^T z_{c_t})},
\end{equation}
where $c_i \in C$ and $k$ is the number of negative samples. Because the vectors are updated in each iteration, the distribution $p_{c_t,c_i}$ is updated during training.

To simulate the learning process from easy to hard, the threshold function $t(\eta)$ is defined to control whether the negative sample should be discarded in the negative sampling process. The smaller the $t(\eta)$, the higher the probability that easy negative samples are obtained. Therefore, $t(\eta)$ should be a low value in the early stage of training to sample easy negative samples and a large value to include harder negative samples,
\begin{equation}\label{equ:filter}
t(\eta)=\max(\frac{4}{B^{2}}\eta^{2}+\frac{1}{B}, 1),
\end{equation}
where $B$ is the batch size and set to $10^{3}$ by default. We use the default parameter in~\cite{GAO:2018:SNE} where $\eta$ is set to $1$ as the initial value and increases by $1$ with every batch. When $p(w|c_t)\geq t(\eta)$, the probability $p_{c_t,w}$ in sampling $w$ is set to zero, indicating that difficult negative samples are discarded. 
In the training process, we gradually include difficult negative samples by increasing $t(\eta)$ and $t(\eta)=1$ means all negative samples are considered. Using self-paced learning, we start with easy negative samples when learning a new model, and then the difficult ones are gradually learned to refine the model. These samples are the most informative and suitable in the training process.

\subsection{Training Process}

\begin{algorithm}[t]
	\caption{voxel2vec}
	\begin{algorithmic}[1]
		\For {each sampled voxel in volume data}
		\State $c_t$ is the scalar value of the voxel;
		\State $O_t$ is the context positive sample set of the voxel;
		
		\For {each $o_{i}\in O_t$}
		\State $g \gets 1 - \sigma (\hat{z}_{o_i}^{T} z_{c_t})$;
		\State $d_1 \gets g\hat{z}_{o_i}^{T} - \lambda \frac{z_{c_t}}{||z_{c_t}||_2} $;
		\State $e \gets d_1\cdot \alpha$;
		\State $d_2 \gets gz_{c_t}$;
		\State $\hat{z}_{o_i} \gets \hat{z}_{o_i} + d_2\cdot \alpha$;
		\State Draw $k$ negative samples from $D = p_{c_t, v}$ where $p_{c_t, v} = 0$ for $v\in \{c_t \cup O_t\}$ or $p(v|c_t)\geq t(\eta)$;
		
		\For {each negative sample $w$}
		\State $g \gets - \sigma (\hat{z}_w^{T} z_{c_t})$;
		\State $d_1 \gets g\hat{z}_w^{T} $;
		\State $e \gets e+d_1\cdot \alpha$;
		\State $d_2 \gets gz_{c_t} - \frac{\lambda}{k+1}\frac{\hat{z}_w}{||\hat{z}_w||_2}$;
		\State $\hat{z}_w \gets \hat{z}_w + d_2\cdot \alpha$;
		\State $p_{c_t,w} \gets \frac{exp(\hat{z}_w^T z_{c_t})}{\sum_{j=1}^{k}exp(\hat{z}_{w_j}^T z_{c_t})}$;
		\EndFor
		\State $z_{c_t} \gets z_{c_t}+e$;
		\EndFor
		
		\EndFor
	\end{algorithmic}
	\label{alg:process}
\end{algorithm}

\begin{figure*}[t]
	\centering
	\includegraphics[width=0.97\textwidth]{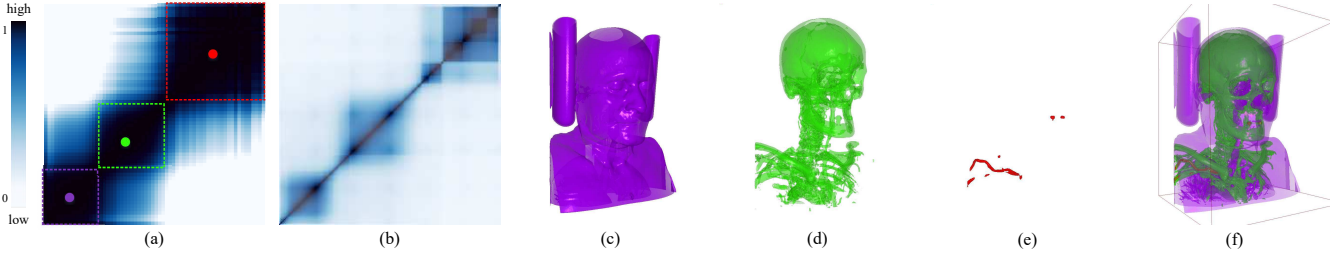}\\
	\caption{
		Similarity analysis of MANIX. Each pixel in (a) and (b) show the similarity between any two scalar values calculated by voxel2vec and isosurface-based similarity~\cite{BRUCKNER:2010:ISM} using the same color map, respectively. A linear color scheme, ranging from light blue to dark blue, is used to encode the similarity value from 0 to 1. Three representative iso-values are selected from the diagonal labeled by three colored points in (a), and their isosurfaces are shown in (c), (d) or (e) with the same color. The three isosurfaces are fuzed in (f).
	}\label{fig:manix}
\end{figure*}

\begin{figure*}[t]
	\centering
	\includegraphics[width=0.97\textwidth]{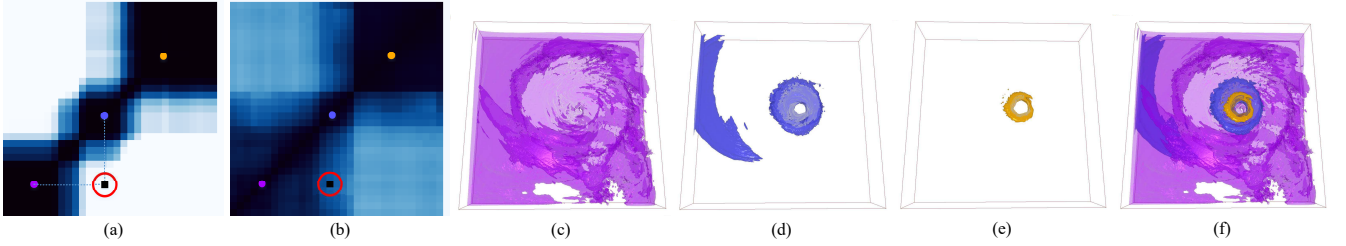}\\
	\caption{Similarity analysis of the speed variable in the hurricane Isabel simulation data set. (a-b) Similarity maps of voxel2vec and isosurface-based similarity~\cite{BRUCKNER:2010:ISM} with the same colormap in Fig.~\ref{fig:manix}, respectively. (c-e) Isosurfaces of selected iso-values in (a) and (b) with the same color. (f) All isosurfaces.
	}\label{fig:univariate_isabel}
\end{figure*}

In the training process, we go through and sample voxels with different context in volume data. For each voxel sampled, we obtain its scalar value $c_t$ and its context scalar value set $O_t$. For each positive sample in $O_t$, $k$ negative samples are drawn from the adaptive and dynamic distribution based on two strategies for negative sampling. A gradient ascending method is used to update the distributed representation $z_c$ and context distributed representation $\hat{z}_c$ for the involved scalar values to maximize the probability of Eq.~\ref{equ:negative-1}. 
The training process of voxel2vec for each sampled voxel is summarized in Algorithm~\ref{alg:process}, and the learning rate is $\alpha = 0.05$ for stable training. In this paper, only one epoch is used in the training process of voxel2vec since there are enough training samples for volumes. For how to choose the number of epochs for voxel2vec, please refer to the supplementary material.

\section{Evaluation on Univariate Data}

Based on the transformation from scalar values to symbols and the skip-gram model with negative sampling, voxel2vec can learn distributed representations to represent complex relationships between scalar values in volume data. In this section, we evaluate voxel2vec on univariate data. Because the spatial distribution of a scalar value can be approximated as its isosurface, the relationship between scalar values, i.e., spatial proximity in voxel2vec, is equivalent to the similarity between isosurfaces~\cite{BRUCKNER:2010:ISM}~\cite{HAIDACHER:2011:VAU}~\cite{TAO:2018:ETM}.

The similarity between scalar values is measured by the cosine similarity between distributed representations as follows:
\begin{equation}
\label{equ:similarity}
sim(x,y)=\max (\frac{z_x^T z_y}{||z_x||_2 ||z_y||_2}, 0),
\end{equation}
where $x$ and $y$ are two scalar values.
Since the cosine value between two vectors may be negative, we assume two scalar values are fully dissimilar if their cosine value is negative and assign a zero similarity value to this scalar value pair. In this way, we can construct a similarity map, i.e., a symmetric matrix, to represent the similarity between scalar values.

\subsection{Comparisons}
\label{sec:comparison}

Our similarity map is semantically similar to the ISM~\cite{BRUCKNER:2010:ISM}~\cite{TAO:2018:ETM}, although the calculation methods are completely different. The ISM calculates the distance distribution from all voxels to an isosurface and considers mutual information between two distance distributions as the similarity of two iso-values. The proposed voxel2vec measures neighboring context similarity, i.e., the co-occurrence relationships among scalar values. We first evaluate our method by comparing the obtained similarity map with the ISM on univariate data.

The CT data set MANIX was used to demonstrate the effectiveness of the ISM in selecting representative isosurfaces~\cite{BRUCKNER:2010:ISM}. We apply voxel2vec to this CT data set, and the similarity map is shown in Fig.~\ref{fig:manix}(a), where three features are easily distinguishable within the dotted boxes. Overall, the similarity map is similar to the ISM~\cite{BRUCKNER:2010:ISM} (Fig.~\ref{fig:manix}(b)) for this data set. 
Fig.~\ref{fig:manix}(c-e) show the three representative iso-values selected from (a), corresponding to the skin, skeleton, and metal, respectively. Fig.~\ref{fig:manix}(f) shows the three transparent features. Note that the calculation time of the isosurface-based method is 231.50s using a GTX 1070 GPU, which is more computationally expensive than voxel2vec (34.16s).

The speed variable of the hurricane Isabel simulation data set is also used for comparison. Fig.~\ref{fig:univariate_isabel}(a) and (b) show the similarity map and the ISM, respectively. 
Overall, the two maps are generally similar to each other. Our similarity map has more low similarities (light blue) than the ISM, making features (blocked dark blue) more easily distinguishable in the similarity map. 
Fig.~\ref{fig:univariate_isabel} (a) shows a low similarity between the scalar values marked in purple and blue, while it is high in (b). However, Fig.~\ref{fig:univariate_isabel} (c) and (d) show no obvious context and spatial similarity between the two isosurfaces. The similarity between two isosurfaces is independently calculated for each scalar-value pair in the ISM, while distributed representations of scalar values are jointly learned in the skip-gram model to preserve the similarity between features globally. The scalar values with less common contexts would be pushed away in vector space, and their similarity value would be low. Three representative iso-values are selected from the similarity map and shown in Fig.~\ref{fig:univariate_isabel}(c-e).

Based on the experiments on both real and simulation univariate data, we can conclude that the similarity map based on voxel2vec is similar to the ISM for univariate data. Since voxel2vec jointly learns the representations of scalar values, it can capture global features better compared to previous methods. This similarity map can be used to guide transfer function design and select representative iso-values.

\subsection{Negative Sampling Strategy Analysis}
\label{sec:optimization_verification}

\begin{figure}[t]
  \centering
  \includegraphics[width=0.95\columnwidth]{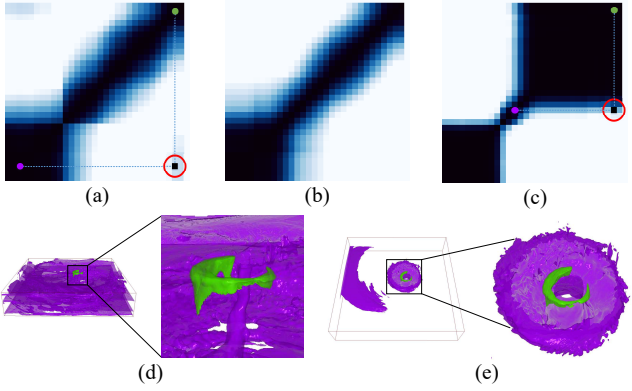}\\
  \caption{The effectiveness of the two strategies on the speed variable in the hurricane Isabel data set. (a-c) Similarity maps generated by voxel2vec with negative sampling without any strategy, with Strategy 1, and with Strategy 2, respectively. Fig.~\ref{fig:univariate_isabel}(a) shows the similarity map with both strategies. (d-e) Spatial distributions of scalar-value pairs marked as black points in (a) and (c), respectively. The colormap is the same as that used in Fig.~\ref{fig:manix}.
  }\label{fig:vertification_1}
\end{figure}

The two strategies are adopted to improve the traditional negative sampling during training. In this section, we demonstrate how these two strategies can enhance the robustness and performance of voxel2vec using the speed variable in the hurricane Isabel simulation data set.

Traditional negative sampling may generate contradictory samples due to the popular neighbor problem. This makes the learning process less robust and leads to vectors with undesirable properties. Fig.~\ref{fig:vertification_1}(a) shows the similarity map with traditional negative sampling. 
We select a scalar value pair, i.e., the black point in the bottom-right region, and present the spatial distributions of the two scalar values in Fig.~\ref{fig:vertification_1}(d). Clearly, these two isosurfaces are not similar in spatial distribution and have no overlap in the context.

Adaptive negative sampling (Strategy 1) can eliminate this type of uncertainty. Fig.~\ref{fig:vertification_1}(b) shows the similarity map trained with only Strategy 1, and the unexpected high similarities in the bottom-right region are reduced. In each training batch, self-paced learning (Strategy 2) selects informative and suitable negative samples from easy to difficult. Generally, scalar values with totally different neighboring contexts are easy negative samples, which indicates that they can be easily classified. However, scalar values with some overlapped neighboring contexts are difficult negative samples, which indicates that they are hard to classify because of a certain degree of similarity. Fig.~\ref{fig:vertification_1}(c) shows the similarity map trained using only Strategy 2. Compared with Fig.~\ref{fig:vertification_1}(a), the unexpected high similarities in the bottom-right region are reduced because they are easy negative samples. There are high similarities in the top-right region in Fig.~\ref{fig:vertification_1}(c) compared to Fig.~\ref{fig:vertification_1}(a). We select one scalar-value pair, which is the black point in the right region, and their spatial distributions are shown in Fig.~\ref{fig:vertification_1}(e). The two isosurfaces are close in the center of the hurricane eye; therefore, they are close in vector space. Finally, the similarity map with clear blocked regions in Fig.~\ref{fig:univariate_isabel}(a) is trained with both strategies, which not only eliminates the popular neighbor problem but also makes the learning process more efficient.
The two strategies are necessary for training a high-quality model for subsequent analysis and visualization.

\section{Applications}

voxel2vec represents scalar values in a low-dimensional vector space to capture complex relationships. This representation can be used to support various volume visualization applications, such as representative iso-value selection for univariate data in Sec~\ref{sec:comparison}. In this section, we describe two applications of voxel2vec: feature classification for multivariate data and association analysis for time-varying and ensemble data.

\subsection{Feature Classification for Multivariate Data}
\label{sec:high_dimension_cluster}

In multivariate data with $N$ variables, the $t$-th voxel has $N$ scalar values $\{s_t^0, s_t^1, ..., s_t^i, ..., s_t^{N-1} \}$, i.e., one scalar-value combination, where $s_t^i$ is the scalar value of the $i$-th variable. Features in multivariate data can be classified by transfer functions~\cite{LJUNG:2016:SOT}, such as 2D transfer functions~\cite{ZHOU:2012:TFC} and high-dimensional transfer function~\cite{GUO:2012:SMV}. The similarity between scalar-value combinations can be used to simplify transfer function design via automatic similarity analysis. If each scalar-value combination can be represented to one point in vector space, these points can be directly clustered to generate features automatically.

We extend the concept of symbols in Sec.~\ref{sec:algorithm} to scalar-value combinations for multivariate data. The symbol that represents each scalar-value combination is unique, i.e., one concatenated word $s_t^0 \_ s_t^1 \_...\_s_t^{N-1}$. 
In this section, we extend voxel2vec to learn the distributed representations of scalar-value combinations and then describe the application on automatic feature classification for two variables and multiple variables via clustering the distributed representations of scalar-value combinations. 
Previous methods~\cite{GUO:2012:SMV}~\cite{LIU:2015:AAF}~\cite{LU:2017:MVD} have focused mostly on identifying representative or significant scalar-value combinations. Our method applies an unsupervised learning model to quantify the relationship/similarity between scalar-value combinations.

\begin{figure}[t]
	\centering
	\includegraphics[width=0.70\columnwidth]{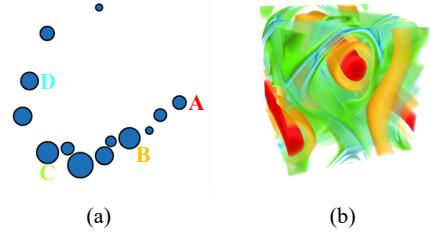}\\
	\caption{
		High-dimensional clustering for the ABC-flow data set. Clusters A, B, C, and D marked in different colors in (a) correspond the spatial distributions of features with the same colors in (b).}
	\label{fig:app2_abcflow}
\end{figure}

\begin{figure*}[t]
	\centering
	\includegraphics[width=1.0\textwidth]{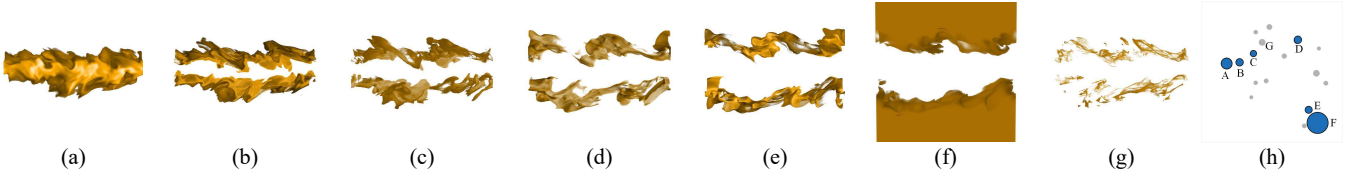} \\
	\caption{High-dimension clustering for the combustion data set. (h) Projection of features via t-SNE. (a-g) Spatial distributions of the features marked by A-G in (h), respectively.}
	\label{fig:app2_multivariate_combustion}
\end{figure*}

\begin{figure*}[t]
	\centering
	\includegraphics[width=1.0\textwidth]{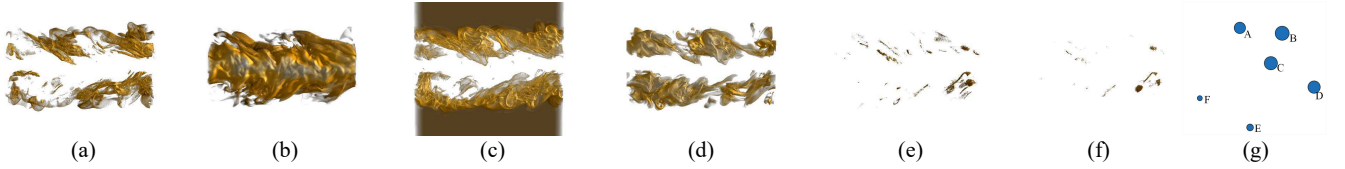} \\
	\caption{Direct clustering of scalar-value combinations for the same variable set in Fig.~\ref{fig:app2_multivariate_combustion} using the Euclidean distance. (g) Projection of features via t-SNE. (a-f) Spatial distributions of the features marked by A-F in (g), respectively.}
	\label{fig:app2_comparison}
\end{figure*}

DBSCAN~\cite{ESTER:1996:ADA} is a clustering algorithm that ignores outlier points and does not require specifying the number of clusters. It is used to cluster scalar-value combinations, where the input space is the corresponding distributed representations. Since voxels with the same scalar-value combinations have the same distributed representations, these voxels in the cluster are grouped as one feature. Then, t-SNE~\cite{JOURNAL:2008:VDU} based on the weighted average vector of the distributed representations of scalar-value combinations in each feature is used to display these features in 2D. The weight is proportional to the number of voxels of each scalar-value combination. The size of each feature is proportional to the number of voxels of the feature with a repulsion force being applied to each feature to avoid occlusion. Thus, the scatter plot enables users to interactively analyze relationships between features. Nearby features may have similar distributed representations of scalar-value combinations, and they may have similar spatial distribution. The parameter $\epsilon$ of DBSCAN is the maximum distance between two points, which can be considered as neighboring points, and $\epsilon = 0.85$ is used in the experiment. {Less interesting features, such as features with less voxels, can be filtered out (Fig.~\ref{fig:app2_abcflow}(a)) or assigned gray color (Fig.~\ref{fig:app2_multivariate_combustion}(h)) interactively.

The clustering of scalar-value combinations is applied to two multivariate data sets to demonstrate the usefulness of the proposed method. The first data is a 3D time-dependent flow, a modified version of the ABC-flow data set~\cite{SAUBER:2006:MAA}, defined as follows:
\begin{equation}\label{equ:time_varying}
v(t)=
\begin{pmatrix}
(A + 0.5t\sin(0.1\pi t))\sin(z)+C\cos(y)
\\ B\sin(x)+(A+0.5\sin(0.1\pi t))\cos(z)
\\  Csin(y)+Bcos(x)
\end{pmatrix},
\end{equation}
where $A=\sqrt3$, $B=\sqrt2$, $C=1$. Eq.~\ref{equ:time_varying} describes an unsteady solution of Euler’s equation. 
Variables $s6$ (average particle velocity magnitude of the path line integrated over $2\pi$ time) and $s7$ (distance between the start and the end point of a path line integration over $2\pi$ time) in the first time step are used for automatic feature classification. As shown in Fig~\ref{fig:app2_abcflow}(a), the features have somewhat of an arc distribution and have different numbers of voxels. Four features along the arc with an even spacing are randomly selected. The relative positions of features reflect their similarity in spatial distribution. For example, the yellow feature (B) lies between the red feature (A) and the green feature (C) in spatial distribution, as shown in Fig~\ref{fig:app2_abcflow}(b).

The second data is three variables from the $51$-st time step of the turbulent combustion data set, i.e., mixture fraction ($\mathrm{mixfrac}$), mass fraction of the hydroxyl radical ($\mathrm{Y\_OH}$), and scalar dissipation rate ($\mathrm{CHI}$). Fig.~\ref{fig:app2_multivariate_combustion}(h) shows seven big features in blue and several small features with few voxels in gray. 
Users can select one feature in the scatter plot and its spatial distribution is visualized in volume rendering. (a-e) show five main features of the combustion from the non-combustion region to the inner layer to the outer layer correspond to features A-E, respectively. 
Feature G is close to feature C in (h), and its size is less than the size of C. As shown in (c) and (g), these two features have similar spatial and numerical distributions, and feature C is larger than feature G. In addition, feature F is the background with the largest number of voxels, and it is close to feature E (outer layer) because they are close in spatial distribution, as shown in (e) and (f).

We further compare our method with the Euclidean distance between scalar-value combinations for the turbulent combustion data set. Except for the distance metric, the clustering algorithm and visual mapping are entirely the same.
Fig.~\ref{fig:app2_comparison} shows the result using the Euclidean distance of scalar-value combinations.
(a-f) are the six spatial features corresponding to A-F in (g). The features in (g) are distributed relatively randomly. 
For example, (a) and (d) are approximately similar in spatial distribution, but they are far away in the projection view. The clustered features are also discontinuous or incomplete, such as (a) and (c). They are unreasonably classified. For example, (b) should be further split into the inner layer and non-combustion region, and (c) should be further split into the background and the outer layer. In contrast, our method can classify scalar-value combinations of multivariate data into high-quality features based on the semantic similarity.

Based on the distributed representations of scalar-value combinations, we can automatically generate features by clustering them in vector space and classify features based on multiple variables. The proposed method is scalable for scalar-value combinations of more than three variables. However, as identified by previous research~\cite{PAR:2004:SCF}~\cite{LU:2017:MVD}, not all variables are relevant, and features can usually be well-defined with several variables rather than with all variables. The set of variables can be selected based on the domain knowledge or the informativeness of variables, such as entropy and mutual information~\cite{BISWAS:2013:AIF}.

\begin{figure*}[thb]
	\centering
	\includegraphics[width=0.99\textwidth]{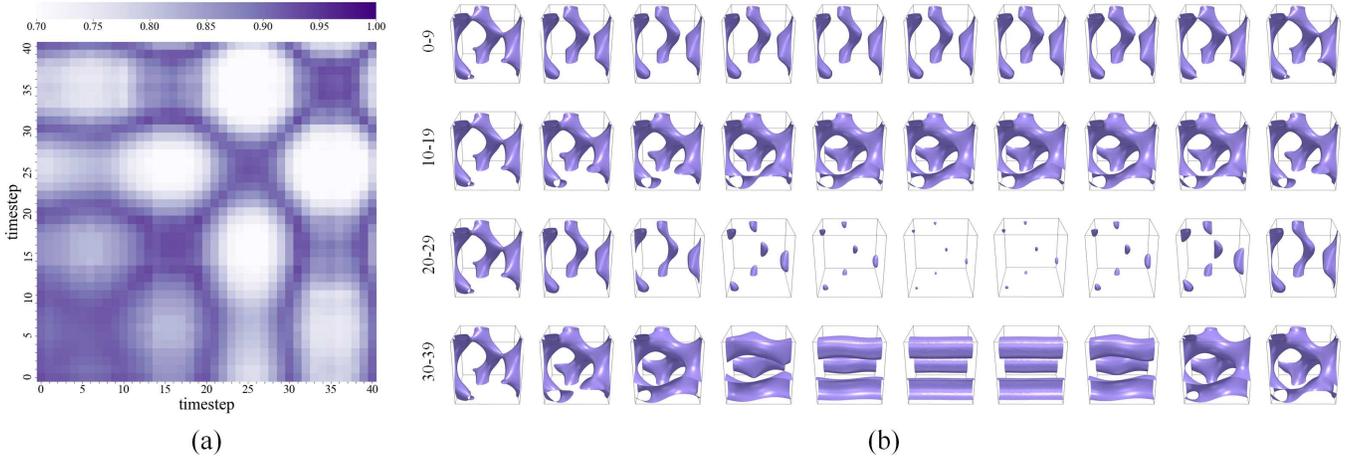}\\
	\caption{Time-varying data association analysis of the flow's velocity magnitude ($s1$) of ABC-flow data set. (a) The association heatmap where the color of each pixel $(i,j)$ presents the association value between $V_i$ and $V_j$. The value from $0.7$ to $1$ reflects in the color bar on the top. (b) The representative isosurfaces (iso-value = $70$) from the $0$-th to $39$-th time steps. The isosurfaces show the spatial variation of volumes over time.}\label{fig:time_varying}
\end{figure*}

\begin{figure}[t]
	\centering
	\includegraphics[width=0.85\columnwidth]{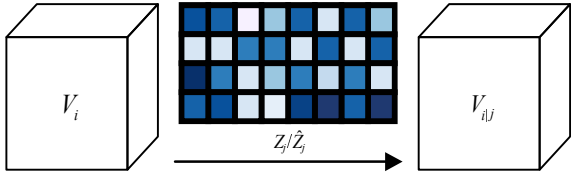}\\
	\caption{After training, voxel2vec learned the model parameters $Z_j$/$\hat{Z}_j$ from volume $V_j$. Transfer prediction uses $Z_j$/$\hat{Z}_j$ to predict the scalar value of each voxel in volume $V_i$ given its context scalar value set, and generates a new volume $V_{i|j}$.}\label{fig:transfer_prediction}
\end{figure}

\subsection{Association Analysis for Time-varying and Ensemble Data}\label{sec:volume_association}

Time-varying data and ensemble data contain many volumes in different time steps and simulation parameters, respectively. These volumes usually have similarities in the numerical and spatial distributions. 
This section investigates the association analysis between volumes in time-varying or ensemble data via voxel2vec. The association between two volumes is quantified by analyzing whether the two volumes have a similar numerical and spatial distributions of scalar values.

voxel2vec learns the context distribution patterns in the volume from the numerical and spatial perspectives. For the $m$-th volume $V_m$ in the time-varying or ensemble data, we can use voxel2vec to learn $Z_m$ and $\hat{Z}_m$. 
Each scalar value after quantization has $z_i \in Z_m$ and $\hat{z}_i \in \hat{Z}_m$ when it is the central scalar value or the context scalar value, respectively. 
In the training, Eq.~\ref{equ:probablity} describes the probability of the context scalar value $o_i$ under the condition of the central scalar value $c_t$. After the training, 
the scalar value of each voxel can be predicted using its context scalar value set $O_t$:
\begin{equation}
P_{O_t}=\frac{1}{|O_t|}\sum_{o_i \in O_t}^{}\sigma(\hat Z^Tz_{o_i}),
\end{equation}
where $P_{O_t} \in \mathbb{R}^R$ is the prediction probability distribution for the central scalar value given $O_t$. For each voxel, we can use its $O_t$ to predict its scalar value, i.e., the scalar value with the highest probability in  $P_{O_t}$ is regarded as the scalar value of the voxel. The predicted scalar value is the most likely result based on the learned voxel2vec model of the volume given $O_t$.} Thus, we can generate a predicted volume $V_{i|i}$ to evaluate the quality of the learned model.

For two volumes $V_i$ and $V_j$ in the time-varying or ensemble data, we can generate the distributed representations $Z_i$/$\hat{Z}_i$ and $Z_j$/$\hat{Z}_j$, respectively. If the numerical and spatial distributions are similar in the volumes $V_i$ and $V_j$, we can reconstruct each voxel in $V_i$ using $Z_j/\hat{Z}_j$ and the context scalar value set $O_t \in V_i$:
\begin{equation}
V_{i|j}(O_t) = \argmax_{s \in [0, R)} P_{O_t}(s),
\end{equation}
where $V_{i|j}$ is the predicted volume. 
In this paper, this task is called transfer prediction, as shown in Fig.~\ref{fig:transfer_prediction}. The similarity between $V_i$ and $V_{i|j}$ indicates whether $ Z_j $/$\hat{Z}_j$ captures the numerical and spatial distributions in $V_i$, i.e., whether the context distribution of scalar values in $V_j$ is similar to the context distribution of scalar values in $V_i$. The similarity between $V_i$ and $V_j$ is quantified by their differences as follows:
\begin{equation}
s(V_i, V_j) = 1-\frac{||V_i-V_j||_{0}}{|V|},
\end{equation}
where $|V|$ is the number of voxels in the volume. We choose the $L_0$ distance to measure how many voxels can be predicted accurately. Since both $V_i$ and $V_j$ can be predicted, the association between the two volumes $V_i$ and $V_j$ is defined as follows:
\begin{equation}\label{equ:association}
ass(V_i, V_j)=\frac{s(V_i, V_{i|j})+s(V_j,V_{j|i})}{2}.
\end{equation}
A higher association value means that two volumes are similar in the context distributions of scalar values. $ass(V_i, V_i)$ is referred to as self-association, indicating the predictability of the volume. A small self-association value means that the context distribution of some scalar values may be inconsistent.

\subsubsection{Time-varying Data}\label{sec:time_varying}

We use transfer prediction to analyze the association between volumes in the ABC-flow data set in Eq.~\ref{equ:time_varying}. The velocity magnitude of the flow $||v(t)||$ ($s1$) is chosen to analyze the relationships between the volumes with a resolution of $256^3$ within the time domain $t \in [0, 40]$.
For any $i$, $j$ $\in [0, 40]$, $V_{i|j}$ and $V_{j|i}$ are generated by the pre-trained $Z_j$/$\hat{Z}_j$ and $Z_i$/$\hat{Z}_i$, respectively, to calculate $ass(V_i, V_j)$. 
The association heatmap presents a symmetric matrix corresponding to the association between all volumes in Fig.~\ref{fig:time_varying}(a). To further verify the effectiveness of the association intuitively, we adopt the iso-value of $70$ as the representative isosurface for each volume from the $0$-th to $39$-th time steps in Fig.~\ref{fig:time_varying}(b).

\begin{figure*}[th]
	\centering
	\subfigure[]{
		\large
		\begin{minipage}[t]{0.92\columnwidth}
			\centering
			\includegraphics[width=\columnwidth]{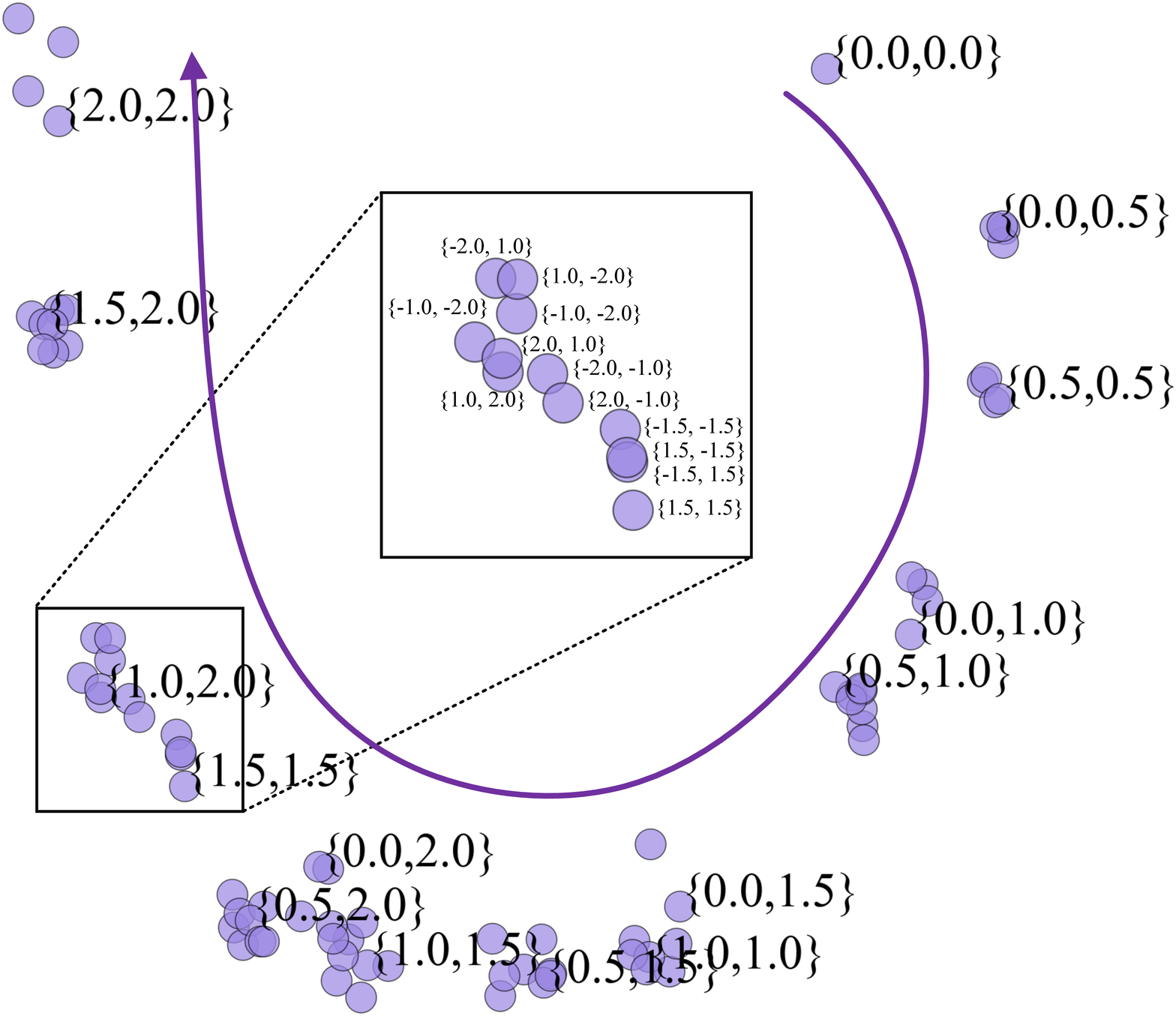}
		\end{minipage}%
	}
	\subfigure[]{
		\large
		\begin{minipage}[t]{0.90\columnwidth}
			\centering
			\includegraphics[width=\columnwidth]{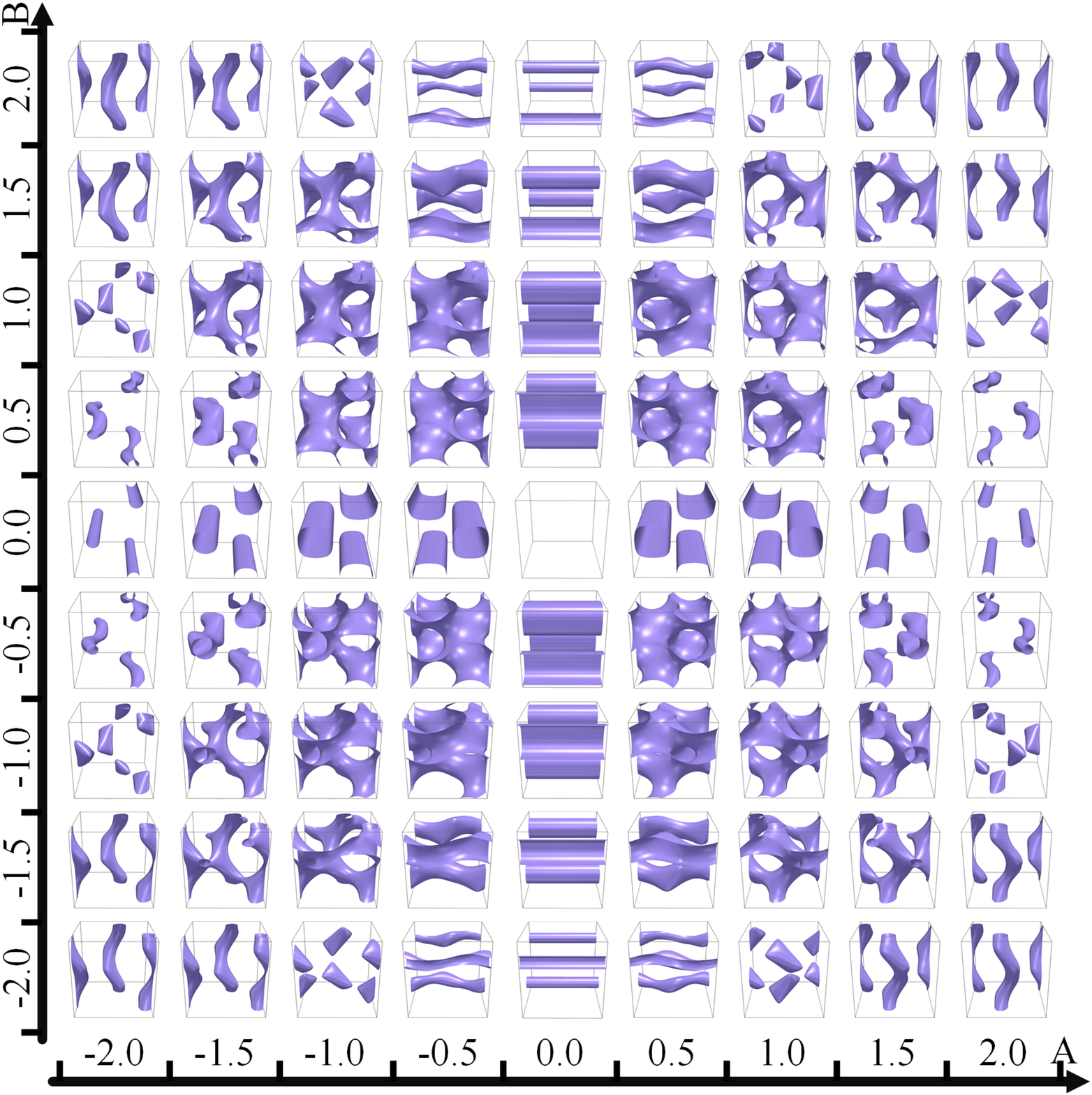}
		\end{minipage}%
	}%
	
	\caption{Ensemble data association analysis of the flow's velocity magnitude ($s1$) of the ABC-flow data set. (a) The scatter plot projected by the matrix $\textit{\textbf{1}}-ass$ of volumes with different parameter combinations of $A$ and $B$. Each purple circle is a parameter combination, and the purple curve shows a clear distribution pattern. (b) The isosurfaces (iso-value=$70$) of the volumes with the parameter of $A$ and $B$, where the horizontal and vertical axes show the parameter $A$ and $B$, respectively.
	}
	\label{fig:ensemble}
\end{figure*}

As shown in Fig.~\ref{fig:time_varying}(a), the association between the volumes shows regular patterns.
First, the mutual prediction of volumes in the time-step set $\{0, 10, 20, 30, 40\}$ shows high associations (marked in yellow). This is reasonable: 
when $t$ is a multiple of $ 10 $, $ ||v(t)|| $ is a constant volume since the sine parts of Eq.~\ref{equ:time_varying} are zero. The isosurfaces corresponding to these volumes also present the same spatial distribution (the first column in Fig.~\ref{fig:time_varying}(b)).
Second, the volumes near about $25$-th time step form a unique feature group in Fig.~\ref{fig:time_varying}(a) (marked in red), represented as an isosurface group (middle of the third row in Fig.~\ref{fig:time_varying}(b)) that is different from other time steps.
Third, the volumes near the $10$-th, $20$-th, $30$-th, and $40$-th time steps change significantly, i.e., low associations with the volumes of adjacent time steps (marked in blue). In contrast, the volumes near the $5$-th, $15$-th, $25$-th, and $35$-th time steps are significantly associated with the volumes of adjacent time steps, as shown in the block area on the diagonal in Fig.~\ref{fig:time_varying}(a).

In this section, voxel2vec demonstrates its ability in detecting approximately periodic changes for time-varying data.
voxel2vec can also identify more general phenomena, such as important events described in the supplementary material.

\subsubsection{Ensemble Data}

We apply transfer prediction to explore similar behavior patterns with multiple parameter combinations in ensemble data.
The ensemble data usually has more than one parameter, causing the association heatmap to fail to show the association under different parameters intuitively. Therefore, we present the association between different parameter combinations in the scatter plot, as shown in Fig.~\ref{fig:ensemble}(a). 
The ABC-flow data set (Eq.~\ref{equ:time_varying}) in the $0$-th time step with two parameters $A $ and $B$ is analyzed in this section, where $A$ and $B$ range from $-2$ to $2$ with a sampling interval of $0.5$ and $C$ is fixed ($C=1$). The number of total parameter combinations is 81.

We calculate the association between volumes in ensemble data. 
$V_p$ represents the volume generated under the parameter combination $p$, such as $p=\{-1.5,2.0\}$, indicating $A=-1.5$ and $B=2.0$. $ass(V_p, :)$ measures the association between the volume under the parameter combination $p$ and those volumes under all parameter combinations. $1-ass(V_i, V_j)$ is the distance between two volumes $V_i$ and $V_j$, and 
t-SNE is used to project all volumes into a 2D space to explore the parameter space of the simulation.

\begin{figure*}[thb]
	\centering
	\subfigure[]{
		\large
		\begin{minipage}[t]{0.99\textwidth}
			\centering
			\includegraphics[width=0.99\textwidth]{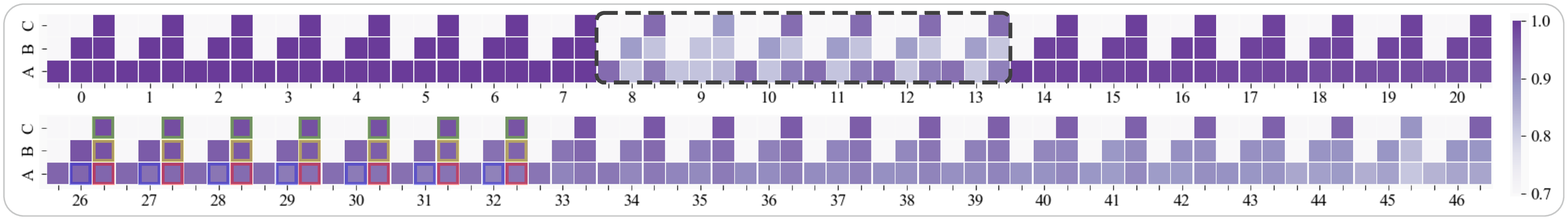}
		\end{minipage}%
		
	}
	
	\subfigure[]{
		\large
		\begin{minipage}[t]{0.99\textwidth}
			\centering
			\includegraphics[width=0.99\textwidth]{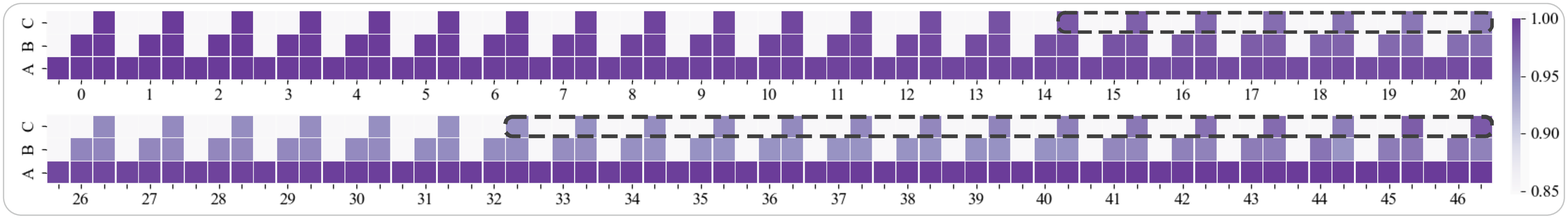}
		\end{minipage}%
		
	}
	\subfigure[]{
		\large
		\begin{minipage}[t]{0.99\textwidth}
			\centering
			\includegraphics[width=0.99\textwidth]{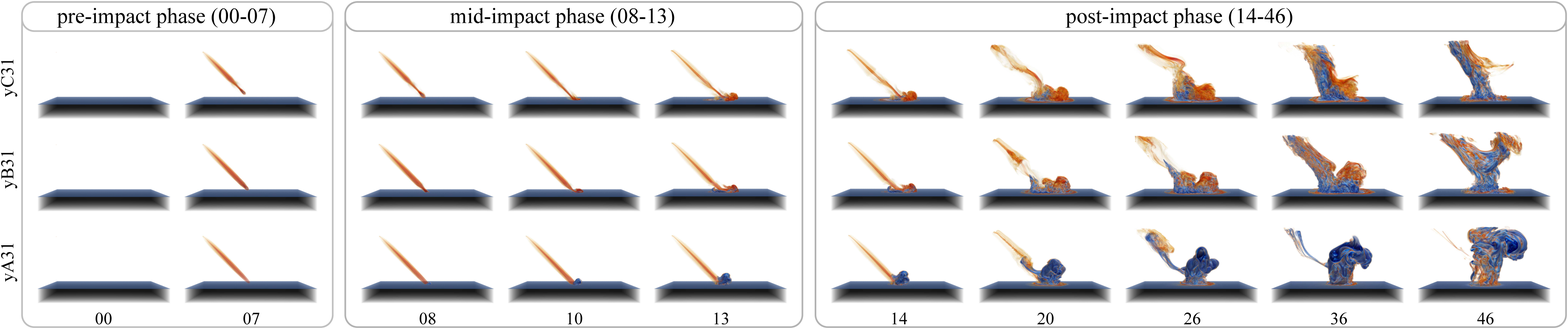}
		\end{minipage}%
		
	}
	\caption{Time-varying and ensemble data association analysis. (a)-(b) The association heatmap of the variables of water fraction ($v02$) and asteroid fraction ($v03$) over time steps, respectively. 
	Each time step corresponds to one association heatmap, and each heatmap encodes the association matrix with three ensemble parameters yA31, yB31 and yC31.
	(c) Ten volumes for each ensemble member. The time steps are divided into three phases: the asteroid's pre-impact, mid-impact, and post-impact.}
	\label{fig:deep_water}
\end{figure*}

The volumes in the ensemble data are projected in the scatter plot (Fig.~\ref{fig:ensemble}(a)) based on their distance matrix $\textbf{\textit{1}}-ass$, where each purple circle in the scatter plot represents a volume under one parameter combination. To avoid visual occlusion, we only label the parameter combinations where $A,B \geq 0$ and $ A\leq B $. 
There are some interesing patterns in Fig.~\ref{fig:ensemble}(a): 
given a positive value pair $(a, b)$, if $|A|=a$ and $|B|=b$, or $|A|=b$ and $|B|=a$, 
the association between the volumes is relatively high, and the corresponding volumes are clustered together, such as the volumes under the parameter combinations $\{1.0, 2.0\}$ and $\{1.5, 1.5\}$ in the middle of Fig.~\ref{fig:ensemble}(a). In this case, the volumes with the $12$ parameter combinations $\{\pm1.0, \pm2.0\}$, $\{\pm2.0, \pm1.0\}$, $\{\pm1.5, \pm1.5\}$ are semantically similar. In other words, the scatter plot intuitively represents the sign's insensitivity and the interchangeability of $A$ and $B$. 
Under this condition, the isosurfaces have a very consistent association in spatial distribution, such as the isosurfaces of the four corners of the value pair $(2.0, 2.0)$ in Fig.~\ref{fig:ensemble}(b). These parameter combinations have similar spatial structures but different spatial positions. 
This is because voxel2vec learns the context distribution patterns of scalar values in the local spatial domain, and not only the spatial distribution or the numerical distribution. 
When $|A|+|B|$ increases linearly, the corresponding parameter combinations are also roughly distributed along the purple curve's direction. The distribution indicates that $|A|+|B|$ is approximately equivalent to one parameter, so the commutation and sign changes of $A$ and $B$ do not affect the overall context distribution of volumes.

\subsubsection{Time-varying Ensemble Data}

The deep water impact ensemble data set with the time domain $t \in [0, 46]$ simulates the entire process of an asteroid's impacts into the sea, where the asteroid is initialized with a 250-meter diameter and a 45-degree momentum. Variables $v02$ (volume fraction of water) and $v03$ (volume fraction of the asteroid) are used for analysis. We choose three ensemble members $yA31$, $yB31$, and $yC31$, representing no airburst as the full asteroid impacts to the sea, an airburst at an elevation of 5 kilometers above the sea level, and at an elevation of 10 kilometers above the sea level, respectively. Note that the volumes from the 21-st to 25-th time steps are not analyzed since those with the ensemble member $yC31$ are not provided on the official website.

For each variable, we calculate the association heatmap of the ensemble members over each time step. Since each heatmap is symmetrical, the heatmaps only show the lower part over each time step in Fig.~\ref{fig:deep_water}(a) and (b). 
The colors from light to dark indicates the association value from low to high. 
Fig.~\ref{fig:deep_water}(c) shows the visualization results of $v02$ and $v03$ in blue and orange, respectively.

Fig.~\ref{fig:deep_water}(a) shows that the association values between the three ensemble members from the $8$-th to $13$-rd time steps are significantly different from those in other periods. 
We divide the entire time step of $v02$ into three phases, namely, the asteroid's pre-impact, mid-impact, and post-impact, and visualize these volumes in Fig.~\ref{fig:deep_water}(c). 
This is consistent with the association heatmap in that the mid-impact phase presents a discontinuous association with adjacent phases, marked in black in Fig.~\ref{fig:deep_water}(a).

For water fraction ($v02$), 
the three phases are analyzed as follows:
\textbf{In the pre-impact phase}, because the asteroid has not yet collided with the sea, the context distributions of $v02$ under the three parameter members are consistent (only the deep water below the sea level), so the ensemble members in this phase have high associations. 
\textbf{In the mid-impact phase}, the asteroid is in contact with the water, and the influence on the context distribution of volumes suddenly increases, manifested as sudden light colors marked in the black region in the heatmap. For $yA31$, the asteroid is directly injected into the water, and part of the asteroid's potential energy is converted into the kinetic energy of the water, causing most of the water molecules to splash out to the surrounding area. The asteroid under $yB31$ and $yC31$ explode in the air, and thus the amount of splashed water is relatively small, making it difficult to learn the neighboring context of the corresponding scalar values. This makes the learned model unable to effectively predict the context distribution with severe fluctuation under other ensemble members. 
\textbf{In the post-impact phase}, the association gradually decreases, and the self-association of $yC31$ is relatively high, such as only the slight variations of the associations with green marks in Fig.~\ref{fig:deep_water}(a). Therefore, we conclude that the context distribution of $yC31$ is relatively consistent in the post-impact phase. The third phase in Fig.~\ref{fig:deep_water}(c) can verify this assumption: the spatial distribution of $v02$ of $yC31$ above the sea level is only characterized by the backflow of many water molecules along the asteroid's trajectory, while the spatial distributions of $yA31$ and $yB31$ are not only characterized by the backflow but also by the water molecules splashing out the surrounding area.
In addition, for most time steps in the post-impact phase in Fig.~\ref{fig:deep_water}(a), the colors between $yB31$ and $yC31$ (marked in yellow) are slightly darker than those between $yA31$ and $yC31$ (marked in red), and the latter are slightly darker than those between $yA31$ and $yB31$ (marked in blue), from which we can reasonably infer that $yB31$ is more associated with $yC31$ than $yA31$.

\begin{table*}[t]
	\centering
	\caption{Information on experiments in each figure, including the data set, quantization level $R$, resolution, variable set, number of symbols $|C|$, and the average training time of voxel2vec for each task (in seconds).}
	\label{tab:efficiency}
	\resizebox{\textwidth}{!}{%
		\begin{tabular}{@{}ccccccccc@{}}
			\toprule
			\textbf{\makecell[c]{Figure(s)}}                                      & \textbf{Data set} & \textbf{Resolution}       & \textbf{\makecell[c]{$R$}} & \textbf{Time step} & \textbf{Variable set} &\textbf{ \makecell[c] {Ensemble \\ parameters}}   & \textbf{ \makecell[c]{ $|C|$}} & \textbf{Time} \\ \midrule
			Fig.~\ref{fig:manix}
			& MANIX          & $256\times 256\times 230$ & 256                        & -                  & -              & -         & 256                    & 34.16         \\
			Fig.~\ref{fig:univariate_isabel}, Fig.~\ref{fig:vertification_1}
			& Hurricane Isabel  & $500\times 500\times 100$ & 32                         & 21                 & \{speed\}      & -         & 32                    & 14.99        \\
			
			Fig.~\ref{fig:app2_abcflow}
			& ABC-flow          & $256\times 256\times 256$ & 32                         & 0                  & \{$s6$, $s7$\}     & -         & 138                   & 44.91        \\
			Fig.~\ref{fig:app2_multivariate_combustion}, Fig.~\ref{fig:app2_comparison}           
			& Combustion        & $480\times 720\times 120$ & 16                          & 51                 & \{mixfrac, Y\_OH, chi\} & - & 588                    & 1452.56        \\
			Fig.~\ref{fig:time_varying}
			& ABC-flow         & $256\times 256\times 256$ & 256                         & 0-40                & \{$s1$\}      & -         & 256                   & 42.12        \\
			Fig.~\ref{fig:ensemble}
			& ABC-flow         & $256\times 256\times 256$ & 256                         & 0                & \{$s1$\}        & {A, B}       & 256                   & 42.12        \\
			Fig.~\ref{fig:deep_water}
			& Deep Water Impact         & $300\times 300\times 300$ & 256                         & 0-46                & \{v02, v03\}      & \{yA31, yB31, yC31\}         & 256                   &  54.67        \\
			\bottomrule
		\end{tabular}%
	}
\end{table*}

For asteroid fraction ($v03$), the asteroid rubs against the air, leaving a volume fraction on the trajectory, as shown in the orange features of the first two phases in Fig.~\ref{fig:deep_water}(c). However, the association of the ensemble parameters $yB31$ and $yC31$ in the post-impact phase gradually decreases (approximately from the $14$-th to the $32$-nd time step). This is because the volume fraction rapidly spreads out around and the numerical distribution violently fluctuates. Thus, the context distribution is difficult to train, resulting in a decrease in prediction accuracy. 
The association gradually increases from the $32$-nd to the $46$-th time step because the asteroid's volume fraction becomes stable at the end of the explosion. 
We mark the association of $yC31$ in the black dotted boxes for the two trends in Fig.~\ref{fig:deep_water}(b). The asteroid of $yA31$ does not explode, so the context distribution of $v03$ is stable during the simulation. 
Fig.~\ref{fig:deep_water}(b) shows that a volume may be more associated with others than itself (e.g., the volume with the ensemble parameter $yB31$/$yC31$ in the 26-th time step). This is due to that the volume with $yB31$/$yC31$ not only contains similar contexts to the volume with $yA31$, but also has many inconsistent or chaotic contexts that are hard to learn.
Fig.~\ref{fig:deep_water}(b) also shows that $yA31$ has high associations with $yB31$ and $yC31$ because there are always dark colors over time, presumably because voxel2vec learned the context distribution of the asteroid's volume fraction in the trajectory for $yB31$ and $yC31$, which is consistent with the context distribution of $yA31$.

\section{Discussion}\label{sec:discussion}

voxel2vec is an unspervised learning model representing scalar values/scalar value combinations as distributed representations using the context relationships between them, whereby the learning processing is performed in preprocessing. Users can interactively analyze the learned vectors of scalar values/scalar-value combinations in feature classification and association analysis applications for univariate data, multivariate data, time-varying data, and ensemble data. This section discusses the performance, some details of voxel2vec, and differences with related research.

\textbf{Hyper-parameter}.
There are a few hyper-parameters in voxel2vec. The context-window size $n$ 
is used to limit the context window of voxels.
The number of negative samples $k$ determines how many negative samples are fed in for each positive sample. A larger $k$ would generate a better model but require more time for the training. The vector dimension $d$ determines the upper limit of the learned information, but a higher $d$ may cause overfitting. The penalty coefficient $\lambda$ is used to 
prevent overfitting. A $\lambda$ between 0.001 and 0.01 performs well on the vast majority of volumes.
The training loss decreases when feeding more data, and using one pass of 50\% of training samples is close to convergence. However, the convergence of training results is not easy to guarantee using one pass of 50\% of training samples due to the randomness of sampling.
The discretization (related to the quantization level $R$) is necessary to ensure sufficient training samples for each scalar value and capture common context information. Although one volume may only have $R$ scalar values, there may be $R^2$ scalar-value combinations for two volumes. We recommend 
setting $n$ to 1, $k$ to 3, $d$ to 30, $\lambda$ to $0.005$, the number of epochs to 1, and $32 \leq R \leq 256$.
For hyper-parameter analysis, please refer to the supplementary material.

\textbf{Performance}.
The experiments were performed on an Intel Core i7 3.40GHz CPU and a NVIDIA GTX 1070 GPU. Detailed information on experimental data sets and the performance of voxel2vec on these data sets are provided in Table~\ref{tab:efficiency}. The training time ranges from several seconds to several minutes, roughly proportional to the number of voxels and scalar values.
The efficiency of voxel2vec is high, as voxel2vec is not a deep neural network model (only one hidden layer) and negative sampling is used to select a small number of negative samples (instead of all samples) for fast training. 
Note that the learning process takes less time than the GPU-accelerated version of the isosurface-based similarity~\cite{IMRE:2017:EGC}~\cite{TAO:2018:ETM} for MANIX in Fig.~\ref{fig:manix}.
Currently, voxel2vec is implemented in C++, and the process can be further accelerated by using GPU-based learning libraries, such as PyTorch~\cite{PASZKE:2017:ADI}.

\textbf{voxel2vec model}.
In scientific data, scalar values usually have an uneven statistical distribution. High-frequency scalar values are trained many times, while those with a low frequency are trained only few times. High-frequency scalar values may correspond to the background or regions of large features, and they can be trained well.
Low-frequency scalar values may be clustered into important features, such as the hurricane eye in the Isabel data set. Still, the distributed representations of these scalar values may be poor due to the lack of positive samples.
To solve this biased distribution problem, in the training process, we dynamically control the sampling frequency of scalar values according to the frequency in volume: when the frequency of a scalar value is high, the sampling would be appropriately reduced, while the sampling frequency would be increased when the corresponding frequency is low. This strategy can effectively reduce the impact of biased distribution and improve the quality of low-frequency scalar values.

Overfitted models make it easy to take advantage of seemingly correct but spurious relationships, which often mislead feature learning. 
In voxel2vec, the model parameters (related to vector dimension) need to be large enough to capture full underlying information. 
However, a large number of model parameters may cause overfitting because models will over-interpret the scalar values with low frequency and learn many fake features that only exist in those samples. 
In fact, these scalar values may be edges, noise, or unrepresentative features. 
Regularization is a widely-used method to solve overfitting in machine learning by introducing additional new information. The information usually takes the form of a penalty for model complexity.
In this paper, norm penalization (Eq.~\ref{equ:negative-1}) is used to improve generalization ability, keep the model simple, and prevent overfitting.

\textbf{Comparison}.
Machine learning has been widely used in scientific visualization, including feature classification~\cite{HAN:2018:FAD}, in-situ visualization~\cite{HE:2020:IDI}, super-resolution generation~\cite{HAN:2020:TTS}, mesh reconstruction~\cite{WANG:2020:DOR}, etc.
There has also been related work carried out about learning potential features in data and expressing them as low-dimensional vectors. For example, Han et al.~\cite{HAN:2018:FAD} presented a self-supervised deep learning model that encodes the numerical and spatial distribution characteristics of streamlines and stream surfaces into latent feature descriptors in the vector field, which is similar to voxel2vec that learns the distributed representations of isosurfaces for univariate data. In their work, the inputs and outputs are the same but the dimensionality is reduced in the hidden layer in order to get a more dense representation. This is different from voxel2vec: its inputs are the neighboring symbols and the output is the symbol itself. Besides, voxel2vec converts symbols to continuous, dense, and distributed representations that encode the context around the symbol rather than the symbol itself.

Isosurface-based similarity~\cite{BRUCKNER:2010:ISM} measures the similarity between scalar values of univariate data by the geometric distance. It depends on the spatial distribution of scalar values. voxel2vec provides a new perspective to quantify the similarity between scalar values. It analyzes the similarity between scalar values in the neighboring context from a statistical point of view. 
In contrast with independently calculating the geometric distance, voxel2vec jointly learns the representations of scalar values and better captures global features. In addition to scalar values of univariate data, voxel2vec can be easily extended to learn the distributed representations of scalar-value combinations in multivariate data and analyze the overall relationships between volumes in time-varying and ensemble data by transfer prediction.

Correlation coefficient and mutual information are common methods to measure the overall association of volumes~\cite{SAUBER:2006:MAA}~\cite{DUTTA:2017:PIG}~\cite{ZHOU:2018:KTS}. However, correlation coefficient only considers the numerical distribution and completely ignores the spatial distribution of scalar values. 
For example, if a sphere and an ellipsoid have the same numerical distribution and volume resolution, the correlation coefficient between the two volumes would be high. 
In addition, mutual information fails to detect the association between the scalar values without overlap in spatial distribution.
For example, if there is no spatial overlap between two spheres, the similarity calculated by mutual information would be very low.
In contrast, voxel2vec learns different distributed representations for scalar values when the scalar values have different contexts. The scalar values with similar context and without overlap in spatial distribution will be gradually pushed together in vector space.
Therefore, voxel2vec provides a novel association method that measures the overall distribution relationship between scalar values by learning the numerical context and spatial information together.

\section{Conclusion and future work}

This paper proposes an unsupervised learning model named voxel2vec to represent scalar values/scalar-value combinations as low-dimensional vectors to capture complex relationships. These vectors can be used to quantify the similarity between scalar values/scalar-value combinations.
The model is evaluated on univariate data and can generate a similar ISM. 
voxel2vec is further applied to multivariate data for feature classification and time-varying ensemble data for association analysis to demonstrate usefulness and effectiveness of the proposed method.

Future work will extend voxel2vec to feature extraction and tracking for time-varying multivariate data. Moreover, 
transfer prediction can generate the probability distribution for each voxel when predicting the volume itself.
This presents a new opportunity to analyze the uncertainty of scalar values for volume data.

\ifCLASSOPTIONcompsoc
  \section*{Acknowledgments}
  This work was supported by National Natural Science Foundation of China (61890954 and 61972343).
\else
  \section*{Acknowledgment}
   This work was supported by National Natural Science Foundation of China (61890954 and 61972343).
\fi


\ifCLASSOPTIONcaptionsoff
  \newpage
\fi



%

\bibliographystyle{IEEEtran}

\bibliography{voxel2vec_tvcg}

\begin{thebibliography}{10}
\providecommand{\url}[1]{#1}
\csname url@samestyle\endcsname
\providecommand{\newblock}{\relax}
\providecommand{\bibinfo}[2]{#2}
\providecommand{\BIBentrySTDinterwordspacing}{\spaceskip=0pt\relax}
\providecommand{\BIBentryALTinterwordstretchfactor}{4}
\providecommand{\BIBentryALTinterwordspacing}{\spaceskip=\fontdimen2\font plus
\BIBentryALTinterwordstretchfactor\fontdimen3\font minus
  \fontdimen4\font\relax}
\providecommand{\BIBforeignlanguage}[2]{{%
\expandafter\ifx\csname l@#1\endcsname\relax
\typeout{** WARNING: IEEEtran.bst: No hyphenation pattern has been}%
\typeout{** loaded for the language `#1'. Using the pattern for}%
\typeout{** the default language instead.}%
\else
\language=\csname l@#1\endcsname
\fi
#2}}
\providecommand{\BIBdecl}{\relax}
\BIBdecl

\bibitem{BURCKNER:2010:RDE}
S.~Bruckner and T.~M{\"o}ller, ``Result-driven exploration of simulation
  parameter spaces for visual effects design,'' \emph{IEEE Transactions on
  Visualization and Computer Graphics}, vol.~16, no.~6, pp. 1468--1476, 2010.

\bibitem{GUO:2012:SMV}
H.~Guo, H.~Xiao, and X.~Yuan, ``Scalable multivariate volume visualization and
  analysis based on dimension projection and parallel coordinates,'' \emph{IEEE
  Transactions on Visualization and Computer Graphics}, vol.~18, no.~9, pp.
  1397--1410, 2012.

\bibitem{LIU:2015:AAF}
X.~Liu and H.-W. Shen, ``Association analysis for visual exploration of
  multivariate scientific data sets,'' \emph{IEEE Transactions on Visualization
  and Computer Graphics}, vol.~22, no.~1, pp. 955--964, 2015.

\bibitem{LU:2017:MVD}
K.~Lu and H.-W. Shen, ``Multivariate volumetric data analysis and visualization
  through bottom-up subspace exploration,'' in \emph{Proceedings of IEEE
  Pacific Visualization Symposium (PacificVis) 2017}, 2017, pp. 141--150.

\bibitem{Inselberg:1985:TPW}
A.~Inselberg, ``The plane with parallel coordinates,'' \emph{The Visual
  Computer}, vol.~1, no.~2, pp. 69--91, 1985.

\bibitem{Guo:2011:MTF}
H.~Guo, H.~Xiao, and X.~Yuan, ``Multi-dimensional transfer function design
  based on flexible dimension projection embedded in parallel coordinates,'' in
  \emph{Proceedings of IEEE Pacific Visualization Symposium (PacificVis) 2011},
  2011, pp. 19--26.

\bibitem{PAR:2004:SCF}
L.~Parsons, E.~Haque, and H.~Liu, ``Subspace clustering for high dimensional
  data: a review,'' \emph{Acm Sigkdd Explorations Newsletter}, vol.~6, no.~1,
  pp. 90--105, 2004.

\bibitem{ZHOU:2018:KTS}
B.~Zhou and Y.-J. Chiang, ``Key time steps selection for large-scale
  time-varying volume datasets using an information-theoretic storyboard,''
  \emph{Computer Graphics Forum}, vol.~37, no.~3, pp. 37--49, 2018.

\bibitem{PORTER:2019:ADL}
W.~P. Porter, Y.~Xing, B.~R. von Ohlen, J.~Han, and C.~Wang, ``A deep learning
  approach to selecting representative time steps for time-varying multivariate
  data,'' in \emph{Proceedings of IEEE Scientific Visualization Conference
  (SciVis) 2019}, 2019, pp. 40--45.

\bibitem{WANG:2013:SPF}
Y.~Wang, H.~Yu, and K.~L. Ma, ``Scalable parallel feature extraction and
  tracking for large time-varying 3d volume data,'' in \emph{Eurographics
  Symposium on Parallel Graphics \& Visualization}, 2013, pp. 17--24.

\bibitem{LUKASCZYK:2017:NTG}
J.~Lukasczyk, G.~Weber, R.~Maciejewski, C.~Garth, and H.~Leitte, ``Nested
  tracking graphs,'' \emph{Computer Graphics Forum}, vol.~36, no.~3, pp.
  12--22, 2017.

\bibitem{LUKASCZYK:2019:DNT}
J.~Lukasczyk, C.~Garth, G.~H. Weber, T.~Biedert, R.~Maciejewski, and H.~Leitte,
  ``Dynamic nested tracking graphs,'' \emph{IEEE Transactions on Visualization
  and Computer Graphics}, vol.~26, no.~1, pp. 249--258, 2019.

\bibitem{LEISTIKOW:2019:AEV}
S.~Leistikow, K.~Huesmann, A.~Fofonov, and L.~Linsen, ``Aggregated ensemble
  views for deep water asteroid impact simulations,'' \emph{IEEE Computer
  Graphics and Applications}, vol.~40, no.~1, pp. 72--81, 2019.

\bibitem{SAUBER:2006:MAA}
N.~Sauber, H.~Theisel, and H.-P. Seidel, ``Multifield-graphs: An approach to
  visualizing correlations in multifield scalar data,'' \emph{IEEE Transactions
  on Visualization and Computer Graphics}, vol.~12, no.~5, pp. 917--924, 2006.

\bibitem{SUKHAREV:2009:CSO}
J.~Sukharev, C.~Wang, K.~L. Ma, and A.~T. Wittenberg, ``Correlation study of
  time-varying multivariate climate data sets,'' in \emph{Proceedings of IEEE
  Pacific Visualization Symposium (PacificVis) 2009}, 2009, pp. 161--168.

\bibitem{BISWAS:2013:AIF}
A.~Biswas, S.~Dutta, H.-W. Shen, and J.~Woodring, ``An information-aware
  framework for exploring multivariate data sets,'' \emph{IEEE Transactions on
  Visualization and Computer Graphics}, vol.~19, no.~12, pp. 2683--2692, 2013.

\bibitem{DUTTA:2017:PIG}
S.~Dutta, X.~Liu, A.~Biswas, H.-W. Shen, and J.-P. Chen, ``Pointwise
  information guided visual analysis of time-varying multi-fields,'' in
  \emph{Proceeding of SIGGRAPH Asia 2017 Symposium on Visualization}, 2017,
  p.~17.

\bibitem{NAGARAJ:2011:AGC}
S.~Nagaraj, V.~Natarajan, and R.~S. Nanjundiah, ``A gradient-based comparison
  measure for visual analysis of multifield data,'' \emph{Computer Graphics
  Forum}, vol.~30, no.~3, pp. 1101--1110, 2011.

\bibitem{SCHNEIDER:2008:ICO}
D.~Schneider, A.~Wiebel, H.~Carr, M.~Hlawitschka, and G.~Scheuermann,
  ``Interactive comparison of scalar fields based on largest contours with
  applications to flow visualization,'' \emph{IEEE Transactions on
  Visualization and Computer Graphics}, vol.~14, no.~6, pp. 1475--1482, 2008.

\bibitem{SCHNEIDER:2013:ICO}
D.~Schneider, C.~Heine, H.~Carr, and G.~Scheuermann, ``Interactive comparison
  of multifield scalar data based on largest contours,'' \emph{Computer Aided
  Geometric Design}, vol.~30, no.~6, pp. 521--528, 2013.

\bibitem{CARR:2014:JCN}
H.~Carr and D.~Duke, ``Joint contour nets,'' \emph{IEEE Transactions on
  Visualization and Computer Graphics}, vol.~20, no.~8, pp. 1100--1113, 2014.

\bibitem{BRUCKNER:2010:ISM}
S.~Bruckner and T.~M{\"o}ller, ``Isosurface similarity maps,'' \emph{Computer
  Graphics Forum}, vol.~29, no.~3, pp. 773--782, 2010.

\bibitem{BENGIO:2013:RLA}
Y.~Bengio, A.~Courville, and P.~Vincent, ``Representation learning: A review
  and new perspectives,'' \emph{IEEE Transactions on Pattern analysis and
  machine intelligence}, vol.~35, no.~8, pp. 1798--1828, 2013.

\bibitem{COLLOBERT:2011:NLP}
R.~Collobert, J.~Weston, L.~Bottou, M.~Karlen, K.~Kavukcuoglu, and P.~Kuksa,
  ``Natural language processing (almost) from scratch,'' \emph{Journal of
  machine learning research}, vol.~12, no.~12, pp. 2493--2537, 2011.

\bibitem{MIKOLOV:2013:EEO}
T.~Mikolov, K.~Chen, G.~Corrado, and J.~Dean, ``Efficient estimation of word
  representations in vector space,'' \emph{arXiv preprint arXiv:1301.3781},
  2013.

\bibitem{MIKOLOV:2013:DRO}
T.~Mikolov, I.~Sutskever, K.~Chen, G.~S. Corrado, and J.~Dean, ``Distributed
  representations of words and phrases and their compositionality,'' in
  \emph{Proceedings of Advances in neural information processing systems},
  2013, pp. 3111--3119.

\bibitem{FUCHS:2009:VOM}
R.~Fuchs and H.~Hauser, ``Visualization of multi-variate scientific data,''
  \emph{Computer Graphics Forum}, vol.~28, no.~6, pp. 1670--1690, 2009.

\bibitem{Wang:2019:VAV}
J.~Wang, S.~Hazarika, C.~Li, and H.-W. Shen, ``Visualization and visual
  analysis of ensemble data: A survey,'' \emph{IEEE Transactions on
  Visualization and Computer Graphics}, vol.~25, no.~9, pp. 2853--2872, 2019.

\bibitem{HE:2019:MSD}
X.~He, Y.~Tao, Q.~Wang, and H.~Lin, ``Multivariate spatial data visualization:
  a survey,'' \emph{Journal of Visualization}, vol.~22, no.~5, pp. 897--912,
  2019.

\bibitem{BAI:2020:TVV}
Z.~Bai, Y.~Tao, and H.~Lin, ``Time-varying volume visualization: a survey,''
  \emph{Journal of Visualization}, pp. 1--17, 2020.

\bibitem{ZHAO:2010:MRA}
X.~Zhao and A.~Kaufman, ``Multi-dimensional reduction and transfer function
  design using parallel coordinates,'' in \emph{Proceedings of International
  Symposium on Volume Graphics}, 2010, pp. 69--76.

\bibitem{TORGERSON:1952:MSI}
W.~S. Torgerson, ``Multidimensional scaling: I. theory and method,''
  \emph{Psychometrika}, vol.~17, no.~4, pp. 401--419, 1952.

\bibitem{WU:2015:WAF}
F.~Wu, G.~Chen, J.~Huang, Y.~Tao, and W.~Chen, ``Easyxplorer: A flexible visual
  exploration approach for multivariate spatial data,'' \emph{Computer Graphics
  Forum}, vol.~34, no.~7, pp. 163--172, 2015.

\bibitem{Zhou:2014:GSS}
L.~Zhou and C.~Hansen, ``Guideme: Slice-guided semiautomatic multivariate
  exploration of volumes,'' \emph{Computer Graphics Forum}, vol.~33, no.~3, pp.
  151--160, 2014.

\bibitem{WANG:2015:MPM}
Z.~Wang, H.-P. Seidel, and T.~Weinkauf, ``Multi-field pattern matching based on
  sparse feature sampling,'' \emph{IEEE Transactions on Visualization and
  Computer Graphics}, vol.~22, no.~1, pp. 807--816, 2015.

\bibitem{HE:2018:ACF}
X.~He, Y.~Tao, Q.~Wang, and H.~Lin, ``A co-analysis framework for exploring
  multivariate scientific data,'' \emph{Visual Informatics}, vol.~2, no.~4, pp.
  254--263, 2018.

\bibitem{BISWAS:2016:VOT}
A.~Biswas, G.~Lin, X.~Liu, and H.-W. Shen, ``Visualization of time-varying
  weather ensembles across multiple resolutions,'' \emph{IEEE Transactions on
  Visualization and Computer Graphics}, vol.~23, no.~1, pp. 841--850, 2016.

\bibitem{FERSTL:2016:TCA}
F.~Ferstl, M.~Kanzler, M.~Rautenhaus, and R.~Westermann, ``Time-hierarchical
  clustering and visualization of weather forecast ensembles,'' \emph{IEEE
  Transactions on Visualization and Computer Graphics}, vol.~23, no.~1, pp.
  831--840, 2016.

\bibitem{MA:2018:AIF}
B.~Ma and A.~Entezari, ``An interactive framework for visualization of weather
  forecast ensembles,'' \emph{IEEE Transactions on Visualization and Computer
  Graphics}, vol.~25, no.~1, pp. 1091--1101, 2018.

\bibitem{WANG:2016:MCE}
J.~Wang, X.~Liu, H.-W. Shen, and G.~Lin, ``Multi-resolution climate ensemble
  parameter analysis with nested parallel coordinates plots,'' \emph{IEEE
  Transactions on Visualization and Computer Graphics}, vol.~23, no.~1, pp.
  81--90, 2016.

\bibitem{BRUCKNER:2010:BEO}
S.~Bruckner and T.~M{\"o}ller, ``Result-driven exploration of simulation
  parameter spaces for visual effects design,'' \emph{IEEE Transactions on
  Visualization and Computer Graphics}, vol.~16, no.~6, pp. 1468--1476, 2010.

\bibitem{TKACHEV:2019:LPM}
G.~Tkachev, S.~Frey, and T.~Ertl, ``Local prediction models for spatiotemporal
  volume visualization,'' \emph{IEEE Transactions on Visualization and Computer
  Graphics}, vol.~27, no.~7, pp. 3091--3108, 2019.

\bibitem{TAO:2018:ETM}
J.~Tao, M.~Imre, C.~Wang, N.~V. Chawla, H.~Guo, G.~Sever, and S.~H. Kim,
  ``Exploring time-varying multivariate volume data using matrix of isosurface
  similarity maps,'' \emph{IEEE Transactions on Visualization and Computer
  Graphics}, vol.~25, no.~1, pp. 1236--1245, 2018.

\bibitem{OBERMAIER:2015:VTA}
H.~Obermaier, K.~Bensema, and K.~I. Joy, ``Visual trends analysis in
  time-varying ensembles,'' \emph{IEEE Transactions on Visualization and
  Computer Graphics}, vol.~22, no.~10, pp. 2331--2342, 2015.

\bibitem{HAN:2020:V2V}
J.~Han, H.~Zheng, Y.~Xing, D.~Z. Chen, and C.~Wang, ``V2v: A deep learning
  approach to variable-to-variable selection and translation for multivariate
  time-varying data,'' \emph{IEEE Transactions on Visualization and Computer
  Graphics}, vol.~27, no.~2, pp. 1290--1300, 2020.

\bibitem{KEHRER:2012:VAV}
J.~Kehrer and H.~Hauser, ``Visualization and visual analysis of multifaceted
  scientific data: A survey,'' \emph{IEEE Transactions on Visualization and
  Computer Graphics}, vol.~19, no.~3, pp. 495--513, 2012.

\bibitem{PEROZZI:2014:DOL}
B.~Perozzi, R.~Al-Rfou, and S.~Skiena, ``Deepwalk: Online learning of social
  representations,'' in \emph{Proceedings of the 20th ACM SIGKDD international
  conference on Knowledge discovery and data mining}, 2014, pp. 701--710.

\bibitem{GROVER:2016:NSF}
A.~Grover and J.~Leskovec, ``node2vec: Scalable feature learning for
  networks,'' in \emph{Proceedings of the 22nd ACM SIGKDD international
  conference on Knowledge discovery and data mining}, 2016, pp. 855--864.

\bibitem{BERGER:2016:CCD}
M.~Berger, K.~McDonough, and L.~M. Seversky, ``cite2vec: Citation-driven
  document exploration via word embeddings,'' \emph{IEEE Transactions on
  Visualization and Computer Graphics}, vol.~23, no.~1, pp. 691--700, 2016.

\bibitem{ZHOU:2018:VAO}
Z.~Zhou, L.~Meng, C.~Tang, Y.~Zhao, Z.~Guo, M.~Hu, and W.~Chen, ``Visual
  abstraction of large scale geospatial origin-destination movement data,''
  \emph{IEEE Transactions on Visualization and Computer Graphics}, vol.~25,
  no.~1, pp. 43--53, 2018.

\bibitem{XU:2018:EEO}
J.~Xu, Y.~Tao, Y.~Yan, and H.~Lin, ``Exploring evolution of dynamic networks
  via diachronic node embeddings,'' \emph{IEEE Transactions on Visualization
  and Computer Graphics}, vol.~26, no.~7, pp. 2387--2402, 2020.

\bibitem{HAN:2018:FAD}
J.~Han, J.~Tao, and C.~Wang, ``Flownet: A deep learning framework for
  clustering and selection of streamlines and stream surfaces,'' \emph{IEEE
  Transactions on Visualization and Computer Graphics}, 2018.

\bibitem{BOJANOWSKI:2017:EWV}
P.~Bojanowski, E.~Grave, A.~Joulin, and T.~Mikolov, ``Enriching word vectors
  with subword information,'' \emph{Transactions of the Association for
  Computational Linguistics}, vol.~5, pp. 135--146, 2017.

\bibitem{ZIRIKLY:2015:NER}
A.~Zirikly and M.~Diab, ``Named entity recognition for arabic social media,''
  in \emph{Proceedings of the 1st Workshop on Vector Space Modeling for Natural
  Language Processing}, 2015, pp. 176--185.

\bibitem{LAI:2015:RCN}
S.~Lai, L.~Xu, K.~Liu, and J.~Zhao, ``Recurrent convolutional neural networks
  for text classification,'' in \emph{Proceedings of the Twenty-ninth AAAI
  conference on artificial intelligence}, 2015.

\bibitem{ZHANG:2014:BPE}
J.~Zhang, S.~Liu, M.~Li, M.~Zhou, and C.~Zong, ``Bilingually-constrained phrase
  embeddings for machine translation,'' in \emph{Proceedings of the 52nd Annual
  Meeting of the Association for Computational Linguistics (Volume 1: Long
  Papers)}, vol.~1, 2014, pp. 111--121.

\bibitem{ARMANDPOUR:2019:RNS}
M.~Armandpour, P.~Ding, J.~Huang, and X.~Hu, ``Robust negative sampling for
  network embedding,'' \emph{Proceedings of the AAAI Conference on Artificial
  Intelligence}, vol.~33, pp. 3191--3198, 07 2019.

\bibitem{FRIEDMAN:2001:TEO}
J.~Friedman, T.~Hastie, and R.~Tibshirani, \emph{The elements of statistical
  learning}.\hskip 1em plus 0.5em minus 0.4em\relax Springer series in
  statistics New York, 2001, vol.~1, no.~10.

\bibitem{BUDA:2018:ASS}
M.~Buda, A.~Maki, and M.~A. Mazurowski, ``A systematic study of the class
  imbalance problem in convolutional neural networks,'' \emph{Neural Networks},
  vol. 106, pp. 249--259, 2018.

\bibitem{KUMAR:2010:SLF}
M.~P. Kumar, B.~Packer, and D.~Koller, ``Self-paced learning for latent
  variable models,'' in \emph{Proceedings of Advances in Neural Information
  Processing Systems}, 2010, pp. 1189--1197.

\bibitem{GAO:2018:SNE}
H.~Gao and H.~Huang, ``Self-paced network embedding,'' in \emph{Proceedings of
  the 24th ACM SIGKDD International Conference on Knowledge Discovery \& Data
  Mining}, 2018, pp. 1406--1415.

\bibitem{HAIDACHER:2011:VAU}
M.~Haidacher, S.~Bruckner, and E.~Gr{\"o}ller, ``Volume analysis using
  multimodal surface similarity,'' \emph{IEEE Transactions on Visualization and
  Computer Graphics}, vol.~17, no.~12, pp. 1969--1978, 2011.

\bibitem{LJUNG:2016:SOT}
P.~Ljung, J.~Kr{\"u}ger, E.~Gr{\"o}ller, M.~Hadwiger, C.~D. Hansen, and
  A.~Ynnerman, ``State of the art in transfer functions for direct volume
  rendering,'' \emph{Computer Graphics Forum}, vol.~35, no.~3, pp. 669--691,
  2016.

\bibitem{ZHOU:2012:TFC}
L.~Zhou, M.~Schott, and C.~Hansen, ``Transfer function combinations,''
  \emph{Computers \& Graphics}, vol.~36, no.~6, pp. 596--606, 2012.

\bibitem{ESTER:1996:ADA}
M.~Ester, H.-P. Kriegel, J.~Sander, and X.~Xu, ``A density-based algorithm for
  discovering clusters in large spatial databases with noise,'' in
  \emph{Proceedings of the Second ACM SIGKDD International Conference on
  Knowledge Discovery and Data Mining}, 1996, pp. 226--231.

\bibitem{JOURNAL:2008:VDU}
L.~v.~d. Maaten and G.~Hinton, ``Visualizing data using t-sne,'' \emph{Journal
  of machine learning research}, vol.~9, no. Nov, pp. 2579--2605, 2008.

\bibitem{IMRE:2017:EGC}
M.~Imre, J.~Tao, and C.~Wang, ``Efficient gpu-accelerated computation of
  isosurface similarity maps,'' in \emph{Proceedings of IEEE Pacific
  Visualization Symposium (PacificVis) 2017}, 2017, pp. 180--184.

\bibitem{PASZKE:2017:ADI}
A.~Paszke, S.~Gross, S.~Chintala, G.~Chanan, E.~Yang, Z.~DeVito, Z.~Lin,
  A.~Desmaison, L.~Antiga, and A.~Lerer, ``Automatic differentiation in
  {PyTorch},'' in \emph{Proceedings of NIPS Autodiff Workshop}, 2017.

\bibitem{HE:2020:IDI}
W.~He, J.~Wang, H.~Guo, K.-C. Wang, H.-W. Shen, M.~Raj, Y.~Nashed, and
  T.~Peterka, ``Insitunet: Deep image synthesis for parameter space exploration
  of ensemble simulations,'' \emph{IEEE Transactions on Visualization and
  Computer Graphics}, vol.~26, no.~1, 2020.

\bibitem{HAN:2020:TTS}
J.~Han and C.~Wang, ``Tsr-tvd: Temporal super-resolution for time-varying data
  analysis and visualization,'' \emph{IEEE Transactions on Visualization and
  Computer Graphics}, vol.~26, no.~1, pp. 205--215, 2020.

\bibitem{WANG:2020:DOR}
Y.~Wang, Z.~Zhong, and J.~Hua, ``{DeepOrganNet: On-the-Fly Reconstruction and
  Visualization of 3D/4D Lung Models from Single-View Projections by Deep
  Deformation Network},'' \emph{IEEE Transactions on Visualization and Computer
  Graphics}, vol.~26, no.~1, pp. 960--970, 2020.

\end{thebibliography}

%
\begin{IEEEbiography}[{\includegraphics[width=1in,height=1.25in,clip,keepaspectratio]{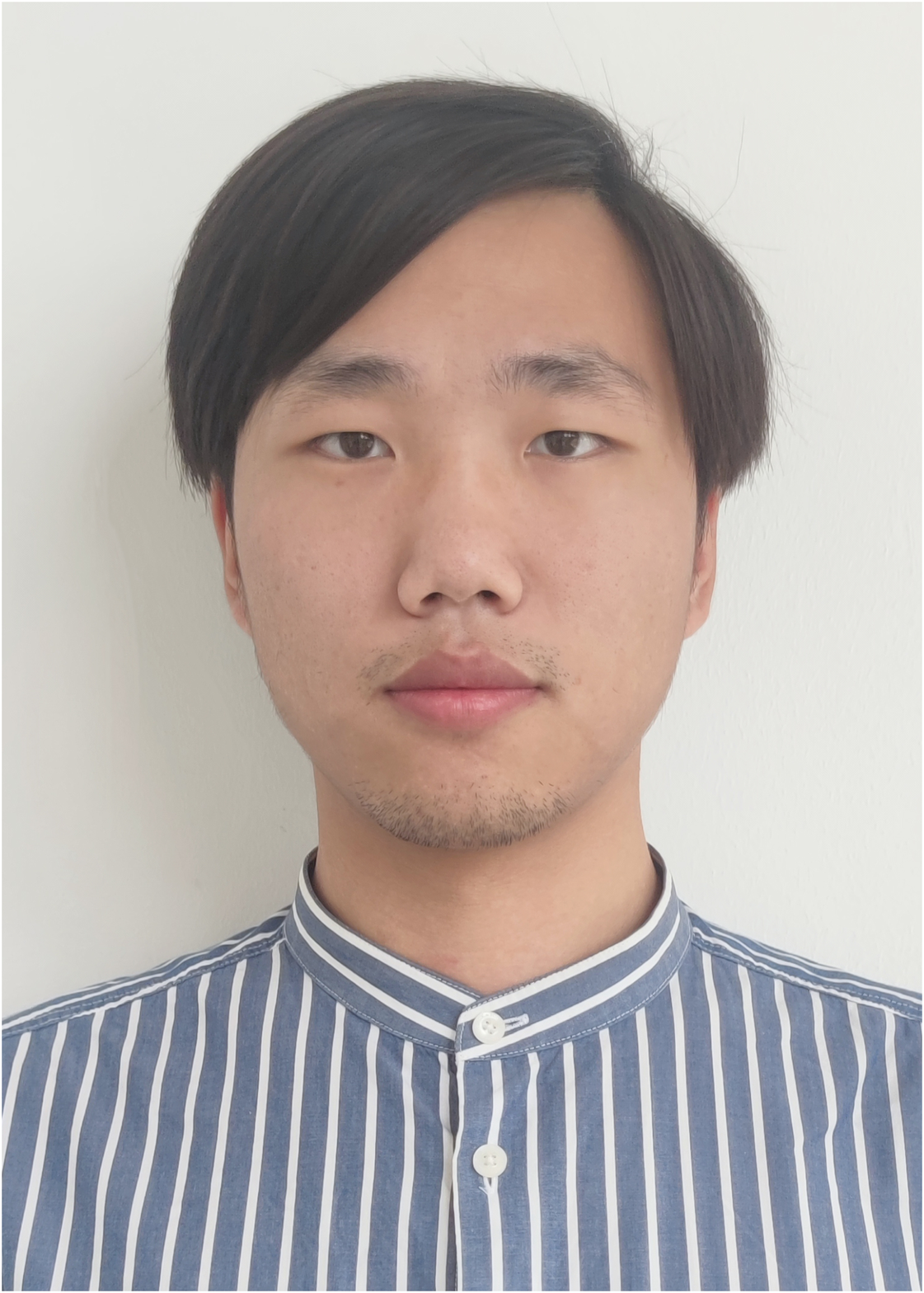}}]{Xiangyang He}
	received the B.S. degree in the College of Software Engineering from Xidian University in 2017. He is currently pursuing the Ph.D. degree at the State Key Laboratory of CAD\&CG in the College of Computer Science and Technology at Zhejiang University. His research interests include scientific visualization and data mining.
\end{IEEEbiography}

\begin{IEEEbiography}[{\includegraphics[width=1in,height=1.25in,clip,keepaspectratio]{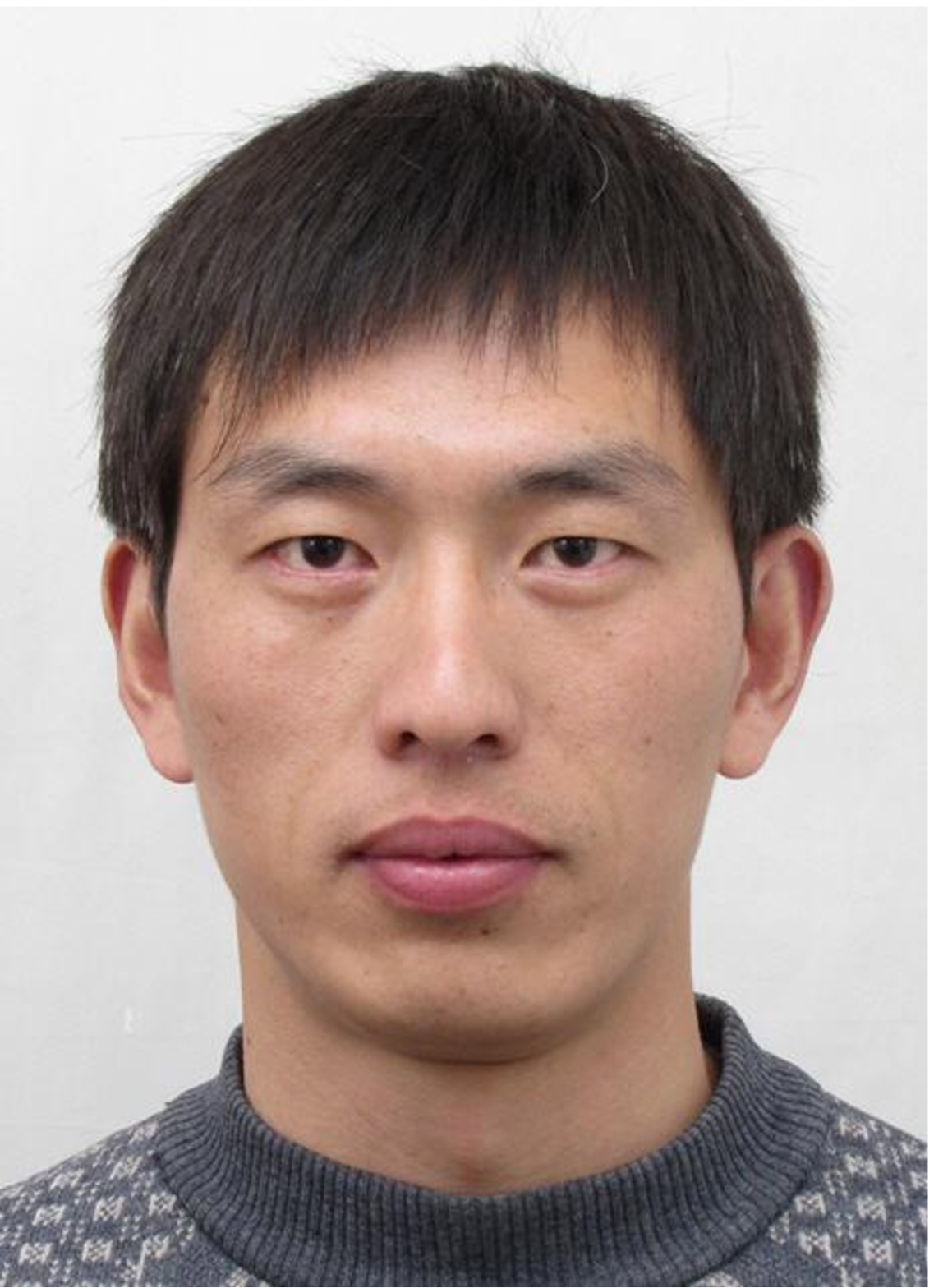}}]{Yubo Tao}
	is an associate professor at the State Key Laboratory of CAD\&CG in the College of Computer Science and Technology at Zhejiang University. He received his B.S. and Ph.D. degrees in computer science and technology from Zhejiang University in China, in 2003 and 2009, respectively. From August 2010 to July 2012, he worked as a Research Fellow in the Centre for Computer Graphics \& Visualization (CCGV) at University of Bedfordshire. His research interests include scientific visualization, visual analytics and computational electromagnetics.

\end{IEEEbiography}

\begin{IEEEbiography}[{\includegraphics[width=1in,height=1.25in,clip,keepaspectratio]{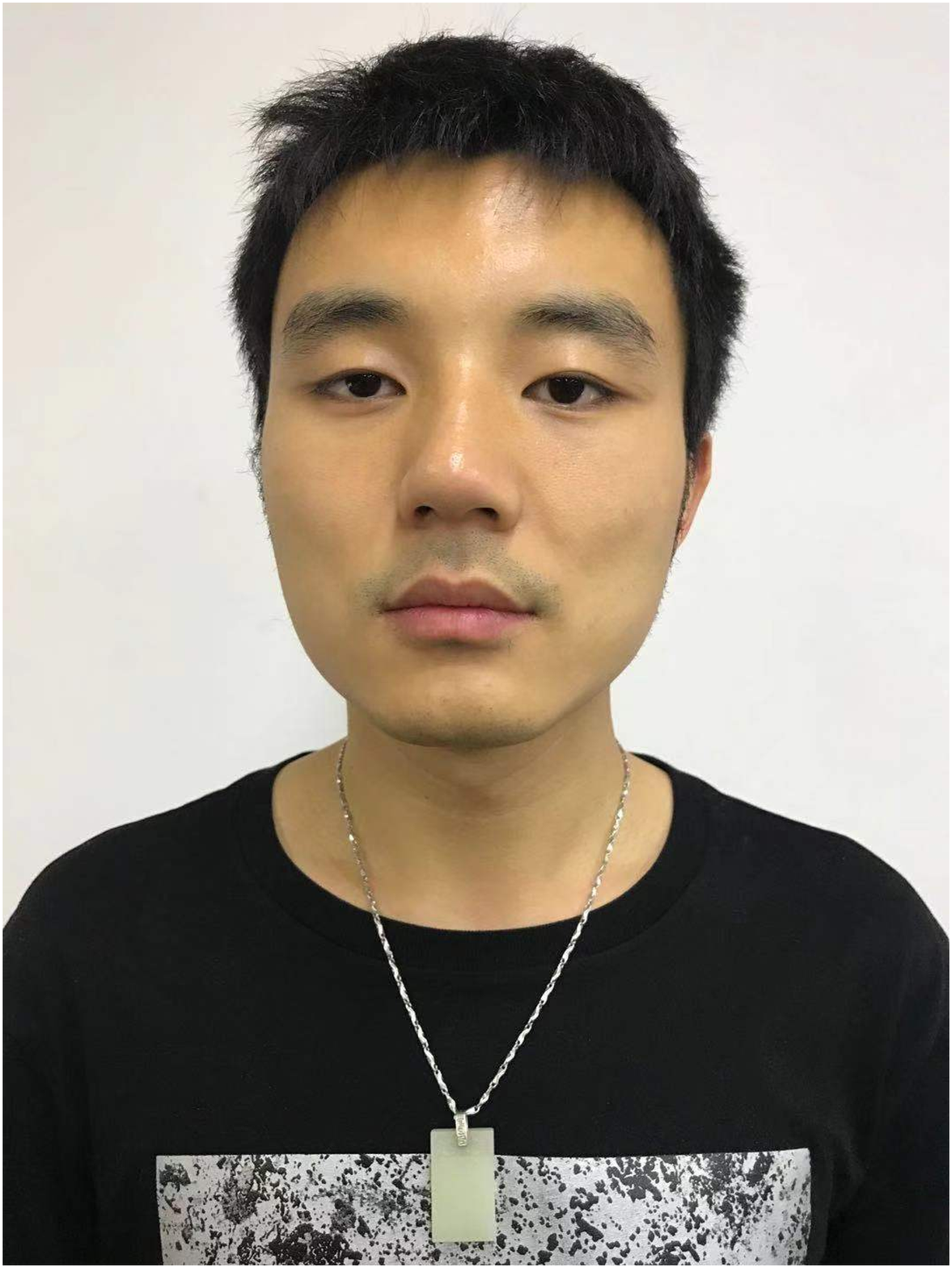}}]{Shuoliu Yang} 
	received his B.S. degree and M.D. degree in the College of Aeronautics and Astronautics at Central South University and the College of Computer Science and Technology at Zhejiang University, in 2017 and 2020, respectively. His research interests include scientific visualization and data mining.
\end{IEEEbiography}

\begin{IEEEbiography}[{\includegraphics[width=1in,height=1.25in,clip,keepaspectratio]{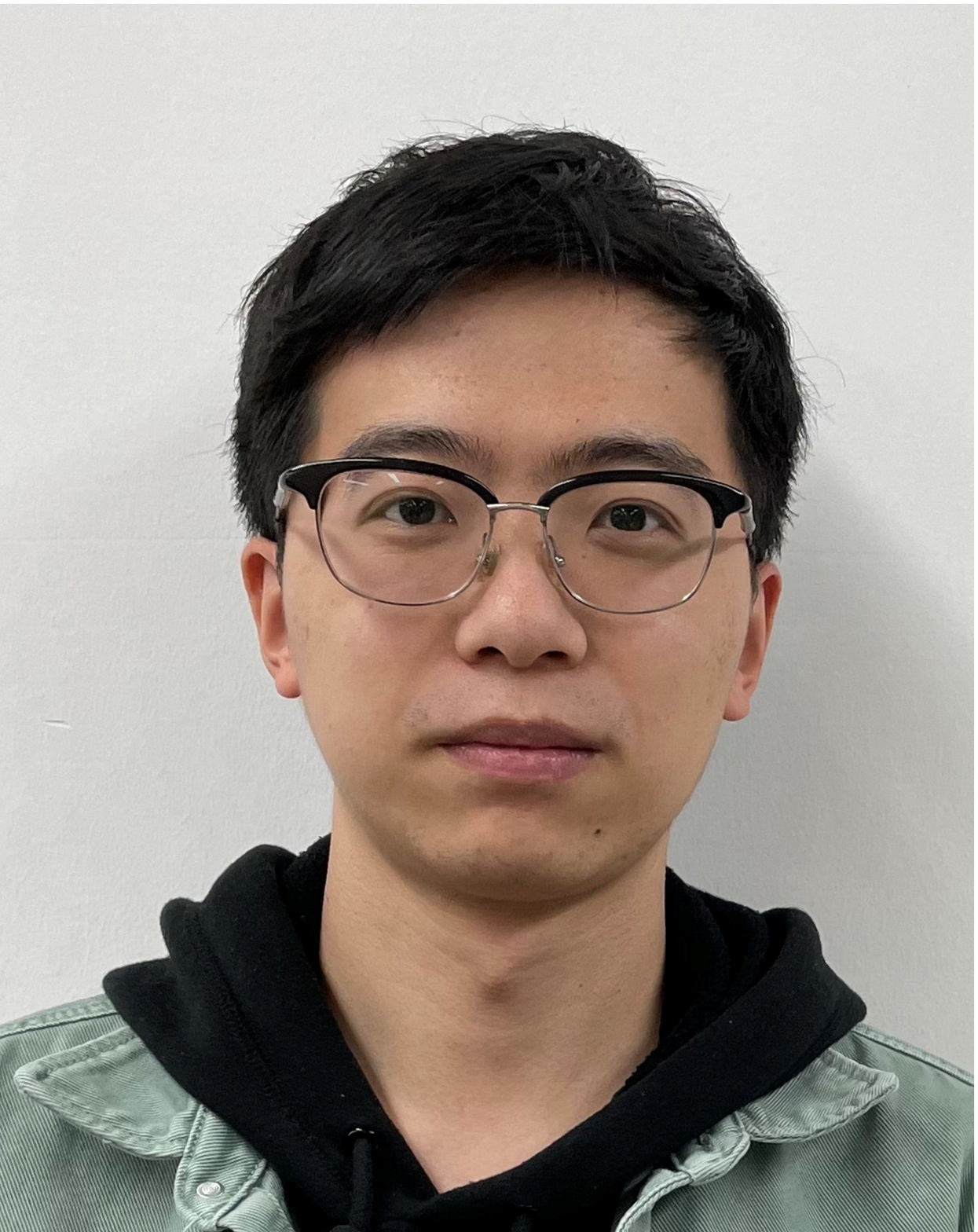}}]{Haoran Dai}
	received his B.S. degree and M.D. degree in the College of Computer Science and Technology at Zhejiang University, in 2019 and 2022, respectively. His research interests include visual analytics and machine learning.
\end{IEEEbiography}

\begin{IEEEbiography}[{\includegraphics[width=1in,height=1.25in,clip,keepaspectratio]{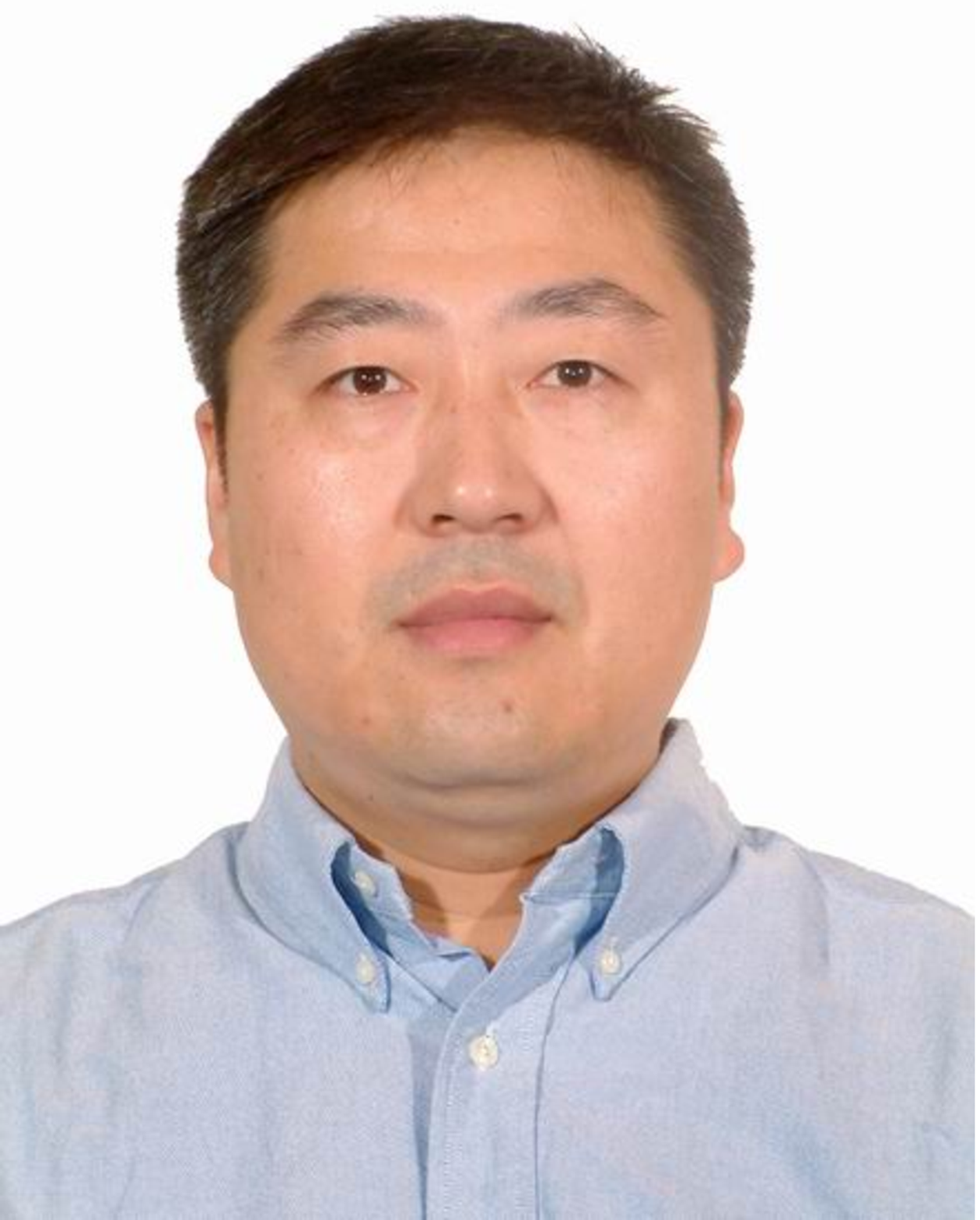}}]{Hai Lin}
	is a professor at the State Key Laboratory of CAD\&CG, Zhejiang University. He received the M.Eng and B.Eng degrees from Xidian University in 1990 and 1987, respectively. He joined the State Key Laboratory of CAD\&CG in 1990. After he obtained a PhD in computer science from Zhejiang University, he had been a Research Fellow in Medical Visualization at	De Montfort University, UK from 2000 to 2003. He was a visiting professor of the Department of Computing and Information Systems, University	of Bedfordshire, UK. His research interests include computer graphics, scientific visualization, volume rendering, virtual reality and graphical electromagnetic computing.
	
\end{IEEEbiography}

\end{document}